\documentclass{article}

\usepackage{arxiv}
\usepackage[utf8]{inputenc}
\usepackage[T1]{fontenc}   
\usepackage{hyperref}      
\usepackage{url}           
\usepackage{booktabs}      
\usepackage{amsfonts}
\usepackage{amsmath}      
\usepackage{nicefrac}      
\usepackage{microtype}     
\usepackage{cleveref}      
\usepackage{graphicx}
\usepackage[square, numbers, sort&compress]{natbib}
\usepackage{doi}
\usepackage{caption}
\usepackage{subcaption}
\usepackage{bm}
\usepackage{algorithm}
\usepackage{algpseudocode}	
\usepackage{setspace}
\usepackage[inkscapepath=svgsubdir]{svg}
\usepackage{tipa}
\usepackage{amssymb}
\usepackage{multirow}
\usepackage{lineno}

\newcommand{\B}[1]{\bm{#1}}

\newcommand{\tran}{^{\mkern-1.5mu\mathsf{T}}}

\newcommand{\fw}{1.0\linewidth}

\newcommand*{\revcolorflag}{black} 
\newcommand*{\rrevcolorflag}{black} 
\newcommand*{\rev}[1]{\textcolor{\revcolorflag}{#1}}
\newcommand*{\rrev}[1]{\textcolor{\rrevcolorflag}{#1}}

\graphicspath{{./figures}}

\title{BINDy -- Bayesian identification of nonlinear dynamics with reversible-jump Markov-chain Monte-Carlo}
\renewcommand{\shorttitle}{BINDy}

\date{}
\author{
	\href{https://orcid.org/0000-0002-3037-7584}{\includegraphics[scale=0.06]{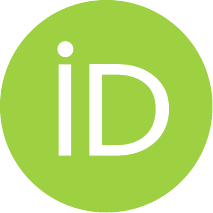}\hspace{1mm}Max D.~Champneys}
	\thanks{Corresponding author} \\
	Dynamics Research Group\\
	University of Sheffield\\
	Mappin St, Sheffield\\
	\texttt{max.champneys@sheffield.ac.uk} \\
	\And
	\href{https://orcid.org/0000-0002-3433-3247}{\includegraphics[scale=0.06]{orcid.pdf}\hspace{1mm}Timothy J.~Rogers} \\
	Dynamics Research Group\\
	University of Sheffield\\
	Mappin St, Sheffield\\
	\texttt{tim.rogers@sheffield.ac.uk} \\
}
\hypersetup{
	pdftitle={Bayesian identification of nonlinear dynamics},
	pdfsubject={Bayesian identification of nonlinear dynamics},
	pdfauthor={Max D Champneys},
	pdfkeywords={Bayesian Statistics, Nonlinear dynamics, Reversible-jump, Markov-chain Monte-Carlo},
}

\begin{document}

\maketitle

\begin{abstract}
	Model parsimony is an important \emph{cognitive bias} in data-driven modelling that aids interpretability and helps to prevent over-fitting. Sparse identification of nonlinear dynamics (SINDy) methods are able to learn sparse representations of complex dynamics directly from data, given a basis of library functions. In this work, a novel Bayesian treatment of dictionary learning system identification, as an alternative to SINDy, is envisaged. The proposed method -- Bayesian identification of nonlinear dynamics (BINDy) -- is distinct from previous approaches in that it targets the full joint posterior distribution over both the terms in the library and their parameterisation in the model. This formulation confers the advantage that an arbitrary prior may be placed over the model structure to produce models that are sparse in the model space rather than in parameter space. Because this posterior is defined over parameter vectors that can change in dimension, the inference cannot be performed by standard techniques. Instead, a Gibbs sampler based on reversible-jump Markov-chain Monte-Carlo is proposed. BINDy is shown to compare favourably to ensemble SINDy in three benchmark case-studies. In particular, it is seen that the proposed method is better able to assign high probability to correct model terms.
\end{abstract}

\keywords{Bayesian Statistics \and Nonlinear dynamics \and Reversible-jump \and MCMC}

\section{Introduction} 

Describing the dynamics of nonlinear processes analytically is of fundamental interest to many branches of scientific modelling. In cases where governing differential equations are unavailable, the practitioner has little choice but to try and discern differential equations directly from data. This difficulty is compounded in situations whereby both the model structure (which terms should be included) and its parameterisation are unknown. In high-noise and low-data regimes, model uncertainty quantification (UQ) becomes critical. Limited data, measurement noise, and an unknown model structure all contribute significant uncertainty.

An important facet of data-driven modelling is model parsimony. Recently, there has been increasing interest in methods that are able to develop parsimonious models directly from measured data. Methods such as sparse identification of nonlinear dynamics (SINDy) \cite{sindy} have been extremely effective at learning sparse representations of complex dynamics. However, a key feature of all SINDy approaches is the specification of heuristic sparsity-inducing methods and hyperparameters that control the inevitable trade-off between fidelity and parsimony in data-driven modelling.

In this paper, the authors take the view that developing parsimonious models from data is inherently an uncertain task. Parsimonious model selection in practice requires interpretable answers to questions in the vein of `what is the probability that a given term should be included given the data?' and `if one were to include this term, how is its value distributed?'. In order to access principled answers to the above, authors propose an alternative to SINDy within a Bayesian framework.

Bayesian uncertainty quantification is a natural framework for conducting UQ in data-driven modelling. Indeed, much has been written on the topic, e.g.\ \cite{schmidt2021probabilistic, cheung2009bayesian}. Physical knowledge and inductive biases (such as parsimony) can be incorporated as \emph{prior knowledge}, making such modelling assumptions explicit. Indeed, many literature contributions propose Bayesian methods for UQ in both model parameterisation \cite{wills2012estimation, fuentes2019efficient, longbottom2024probabilistic} and model structure \cite{wasserman2000bayesian,abdessalem2018model, abdessalem2019model, tripura2023sparse}. Further discussion of specific related works is presented in Section \ref{sec:related}.

\subsection{Contribution}

In this work, interpretable prior distributions are placed over terms in the model (independently of their parameterisation) and over parameters (independently of their inclusion). This prior structure invokes a joint posterior distribution that can be marginalised to obtain useful distributions, that can robustly address modelling questions such as those introduced above.

The key contribution of this work is a novel Bayesian approach to the identification of nonlinear dynamics from a library of basis terms. In particular,

\begin{itemize}
	\item Heuristic sparsity-inducing regression is replaced with interpretable prior distributions over models -- all modelling assumptions are made \emph{up-front}. 
	\item An efficient sampler is proposed to produce samples from the target joint posterior distribution over model terms and parameters.
	\item The proposed approach is shown to compare favourably to ensemble-SINDy in three case studies including a popular population dynamics dataset consisting of only 21 data points.
\end{itemize}

\rev{The remainder of this paper is structured as follows: The following section introduces the necessary background, summarising the SINDy methodology, holding some discussion on parsimony in data-driven modelling motivating the use of reversible-jump Monte-Carlo. A third section introduces the proposed approach and its position in the literature, providing a framework for Bayesian identification of nonlinear dynamics. A fourth section compares the proposed approach to the ensemble SINDy method in three benchmark case studies. A final section presents some discussion and directions for future work.}

\section{\rev{Background}}

\subsection{SINDy methods}

The SINDy method proposed, in 2016 by Brunton et al. \cite{sindy}, has received considerable attention in the scientific literature as a computationally-efficient way to learn differential equations directly from data. Since it was first proposed, the method has been proven to be effective in correctly identifying governing equations from both simulated and real-world data.

At their core, SINDy methods make the key assumption that the observed dynamics admit a first-order formulation that is linear in the parameters of a library of user-specified basis functions $\Theta$. Let $f$ denote the first-order ordinary differential equation (ODE) that describes the evolution of some state-space dynamics $x(t)$,\footnote{\rev{The method can be extended to systems of ODEs, PDEs etc. but for notational simplicity, the exposition in this paper will consider scalar valued ODEs. The authors note that the extension to systems of ODEs (as seen in the case-studies of this manuscript) can be handled by treating each of the regression targets $\dot{x}_i(t)$ independently, from a shared state space comprised of all $x_i$,}}

\begin{equation}
	\dot{x}(t) =  f(x(t))\rev{,}
\end{equation}

where the overdot represents a time derivative. Now the foundational assumption of the SINDy method can be stated,

\begin{equation}
	\dot{x} \approx \Theta(x) \Xi\rev{,}
	\label{eqn:sindy}
\end{equation}

\rev{where $x$ is the measured state, $\Theta(x) \in \mathbb{R}^{N \times D}$ is a library of $D$ user-selected basis functions computed from $x$. In the case that $x$ is a vector, the library of terms may also include functions of all elements in $x$ and cross terms. Thus, it is expected that $D$ will grow much faster than the dimension of the ODE under investigation.} Critically, the coefficient vector $\Xi$ is assumed to be \emph{sparse}, whereby many elements of $\Xi$ are equal to 0. In practice, it is not often the case that measurements of both $x$ and $\dot{x}$ are available. \rev{It is usual in the SINDy literature to assume that only the states, $x$, are measured, and their derivatives $\dot{x}$ must be computed numerically (with sufficient accuracy such that \eqref{eqn:sindy} holds) \cite{sindy}.} The SINDy framework (for identification of ODEs) can thus be summarised by three steps:

\begin{enumerate}
	\item The selection of a library of candidate basis terms.
	\item The selection of a numerical differentiation scheme to compute $\dot{x}$.
	\item Sparse regression to the coefficient vector $\Xi$ by heuristic means.
\end{enumerate}

It is clear that a great deal of methods can be applied to each of the steps above. In this way, SINDy-type methods have come to encompass a family of approaches. Since the original SINDy algorithm was introduced in \cite{sindy}, extensions have been proposed to partial differential equation (PDE) systems \cite{sindy-partial}, implicit dynamics \cite{sindy-pi}, discrete dynamics and weak PDE solutions \cite{sindy-weak}. Furthermore, many methods have been applied to the sparse regression task including sequentially-thresholded least-squares \cite{sindy}, sparse relaxed regression \cite{zheng2018unified} and forward orthogonal least-squares regression (FROLS) \cite{billings2013nonlinear}.
Several approaches also include uncertainty quantification such as the ensemble formulation in \cite{sindy-ensemble} and sparsity-inducing Bayesian methods \cite{hirsh2022sparsifying,tripura2023sparse, fung2024rapid}. Many of these methods have been made readily available to practitioners via an open-source python library \cite{de2020pysindy}.

\rev{Of the three components in SINDy methods described here, this paper is concerned only with the third---determination of appropriate model terms and their parameters. At this stage, it is useful to present some discussion as to the role of parsimony in data-driven modelling.}

\subsection{Parsimony in data-driven modelling}

Inherently, model parsimony is a \emph{cognitive bias}, injected by modellers to promote interpretability and to prevent overfitting. In many sparsity-promoting methods (such as SINDy), parsimony is enforced by proxy; the importance of each term in the library is related to the corresponding size of the parameter in $\Xi$. Although this is a convenient proxy for parsimony, it is not without limitations. A critical limitation in practice is that small parameters in $\Xi$ do not necessarily correspond to small effects in the dynamics. Consider the effect of neglecting a small but negative damping term in the model leading to instability in extrapolation. Practitioners cannot know in advance how small parameters will behave in regions of the state that are not observed in the training data. Normalisation of the columns of $\Theta$ can help to address this shortcoming, but can also introduce ambiguity between correlated library functions. For a motivating example consider the problem of selecting between the terms $x^2$ and $x^4$, $x\in[0,1]$ in the presence of noise. If all terms in $\Theta(x)$ normalised to unit standard deviation, the difference between the two terms might appear trivial in the regime of the training data; the difference in extrapolation is evident. Introducing correlations of this form can ultimately harm model parsimony.

\rev{A particular issue with parameter size as a proxy for term importance is the selection of a threshold value, below which model terms should be excluded. In practice, this level can be difficult to select in a principled manner, leading to a spectrum of models at different levels of sparsity. In \cite{mangan2017model}, an approach based on the Akaike information criterion is used to establish a Pareto front of models trading off error against the number of identified terms. This approach is successful in that the user is offered an interpretable choice from a small subset of possible models. However, the theoretical interpretation of this Pareto front is limited and distributional estimates are not explicitly available. The method also relies on parameter size as a proxy for term importance which can lead to the issues described above.}

Alternative proxies for term inclusion have been considered based on greedy reduction of some error metric. A key example is forward regression least-squares optimisation (FROLS) \cite{billings2013nonlinear, de2020pysindy} that attempts to greedily add (or remove) model terms that maximize an error-reduction ratio. Such approaches circumvent problems of small parameter values described above. In practice however, these methods require the specification of a convergence threshold (or directly number the number of terms in the model, as in the implementation in \cite{de2020pysindy}) which can also be difficult to set in advance in a principled manner.

It is the position of the authors that data alone cannot inform model parsimony. In order to select models that are parsimonious, cognitive biases must be applied.  In this work, a wider probabilistic view of parsimony in SINDy-type methods is taken. Rather than use parameter magnitude as a proxy for sparsity, the authors propose to target the \emph{parameter inclusion probability} directly. Taking a Bayesian view of \eqref{eqn:sindy} and assuming an additive Gaussian noise model one has,

\begin{equation}
	\rev{\dot{x} =  \Theta(x) \Xi  + \epsilon, \quad \epsilon \sim \mathcal{N}(0, \sigma^2),}
	\label{eqn:sindy_bayes}
\end{equation}

where $\sigma^2$ is the noise variance in the observation model. The model identification task at hand is to identify a subset of the columns of $\Theta(x)$. Intuitively, this model set includes all subsets of library functions (the powerset) and thus all possible ways that the parameter vector may be sparsified. \rrev{Let}

\begin{equation}
	\mathcal{M} = \{m_i\}_{i=1}^{2^n}\rev{,}
\end{equation}

be the set of all such possible models for $\Theta(x)$ consisting of $n$ library functions\footnote{The method exposed in this work does not require $n$ to be finite and can be generalised to the case of infinite libraries (for example all polynomials), however the exposition and results herein consider finite $n$ only. Some additional discussion is held in Section 5.}. Key to the formulation here, is that each model in $m\in\mathcal{M}$ is parametrised by a corresponding parameter vector $\Xi$, the dimension of which depends on the model $m$.

The objective of this work is to infer the joint posterior distribution $p(\Xi, m | \dot{x}, \Theta(x))$. The observation model in \eqref{eqn:sindy_bayes} gives rise to a Gaussian likelihood $p(\Xi, m |  \dot{x}, \Theta(x))$ for each model $m$ and corresponding parameter vector $\Xi$. With an appropriately defined prior $p(\Xi, m)$, and applying Bayes rule, one may write,

\begin{equation}
	p(\Xi, m |  \dot{x}, \Theta(x)) = \frac{p(\Xi, m)p(\dot{x} | \Xi, m, \Theta(x))}{p(\dot{x})}\rev{.}
\end{equation}

For notational simplicity in the exposition that follows, the dependence of the likelihood and posterior on $\Theta(x)$ will be dropped, and the observed (or computed) state derivatives are denoted by $\mathcal{D} = \dot{x}$. The above can thus be written,

\begin{equation}
	p(\Xi, m |  \dot{x}, \mathcal{D}) = \frac{p(\Xi, m)p(\mathcal{D} | \Xi, m)}{p(\mathcal{D})}\rev{.}
\end{equation}

Access to the posterior distribution over both the model terms and its parameterisation confers a number of advantages for the data-driven discovery of nonlinear dynamics. If samples from the posterior are available, a major advantage is that they can be marginalised (in the Monte-Carlo sense) to give direct access to a posterior distribution over models, independently of their parameterisation i.e. $p(m | \mathcal{D})$. This allows the practitioner to evaluate the \emph{probability} that a term should be included in a model (given the observed data). This permits more robust ways to introduce model parsimony, for example by excluding terms with an inclusion probability below a certain threshold. Alternatively, the full posterior over model terms and parameters can be propagated to further analyses.

The joint posterior distribution $p(\Xi, m | \dot{x}, \mathcal{D})$ over both models and their parameters is a very challenging object to approach. As is usual in complex Bayesian inference tasks, the model evidence term $p(\mathcal{D})$ is unavailable analytically. However, this is not the only obstacle. The major inferential challenge is that the \emph{dimension} of the parameter vector $\Xi$ has come to depend on the particular model it parameterises. In order to overcome these difficulties, a sampling scheme based on reversible-jump Markov-chain Monte-Carlo (RJMCMC) \cite{green1995reversible} will be employed.

\subsection{Reversible-jump MCMC}

It will be useful here to briefly review both Metropolis-Hastings (MH) and reversible-jump Markov-chain Monte-Carlo theory in a general setting in order to motivate the proposed approach. Much of the exposition here is available in additional detail in references \cite{geyer1992practical} for MH and \cite{green1995reversible} for reversible-jump MCMC. Consider the well-studied problem of sampling from a target distribution $\pi(\theta)$ (that is known up to some constant), by constructing a Markov chain with $\pi$ as its stationary distribution. A sufficient condition for convergence is that of \emph{detailed balance}, whereby,

\begin{equation}
	\pi(\theta')\kappa(\theta | \theta') = \pi(\theta)\kappa(\theta' | \theta)\rev{.}
\end{equation}

Where $\kappa(\theta' | \theta)$ is the kernel of the Markov chain moving from state $\theta$ to new location $\theta'$. The MH algorithm further divides this kernel into a transition density $p(\theta' | \theta)$ and an acceptance probability $\alpha(\theta \rightarrow \theta')$,

\begin{equation}
	k(\theta' | \theta) = p(\theta' | \theta)\alpha(\theta \rightarrow \theta')\rrev{,}
\end{equation}

\rrev{the} familiar MH acceptance probability can be found by maximising the acceptance probability while retaining the condition of detailed balance. The optimal choice is found to be,

\begin{equation}
	\alpha(\theta \rightarrow \theta')  = \min \Big\{ 1, \\
	\underbrace{\frac{ \pi(\theta)}{\pi(\theta)}}_{\substack{\text{Target} \\ \text{ratio}}}
	\underbrace{\frac{ p(\theta | \theta')}{p(\theta' | \theta)}}_{\substack{\text{Proposal} \\ \text{ratio}}}
	\Big\}\rev{,}
\end{equation}

whereby an unknown normalisation constant in $\pi$ can be cancelled from the numerator and denominator.

In the case that $\theta'$ and $\theta$ have different dimensions, detailed balance will not hold for standard choices of the transition density. To overcome this issue, in 1995 Green introduced the reversible-jump MCMC (RJMCMC) algorithm \cite{green1995reversible} as a method to sample from distributions defined over parameters of different dimension. The RJMCMC method overcomes this difficulty via \emph{dimension matching}. Let $k$ be the dimension of the state of a Markov chain at $\theta$. In order to move to a new state with dimension $k'$ and position $\theta'$, an auxiliary variable $u'$ with dimension $j'$ is sampled. To compensate for the mismatch between dimensions, it is required that $j+k = j'+k'$, where $j$ is the current dimension of the auxiliary variable $u$. Then, a bijective map $g_{k \rightarrow k'}: \{u, \theta\} \rightarrow \{u', \theta'\}$ between each pair of dimensions is defined, such that the dimension of $\{u, \theta\}$ is unchanged by the bijection. If the probability of `jumping' from one dimension to another (via the appropriate corresponding bijection) is given by $J(k', \theta' | k, \theta)q(u' | \theta')$ Then the RJMCMC acceptance probability can be reformulated as,

\begin{equation}
	\alpha(\theta \rightarrow \theta')  = \min \Big\{ 1, \\
	\underbrace{\frac{ \pi(\theta')}{\pi(\theta)}}_{\substack{\text{Target} \\ \text{ratio}}}
	\underbrace{\frac{J(k', \theta' | k, \theta)}{J(k, \theta | k', \theta')} }_\text{Jump ratio}
	\underbrace{\frac{q(u' | \theta')}{q(u | \theta)}}_{\substack{\text{Auxillary} \\ \text{ratio}}}
	\underbrace{\left| \frac{\partial g_{k \rightarrow k'}}{\partial (u, \theta)} \right| }_{\substack{\text{Jacobian} \\ \text{determinant}}}
	\Big\}\rev{.}
\end{equation}

In the general setting, working with RJMCMC can be cumbersome for the practitioner. The specification of the bijective maps is non-trivial and has a strong effect on the sampling efficiency of the scheme. Fortunately, a significant simplification is sometimes available and will be employed here. The trick is to sample model parameters independently between models. \rev{This is achieved by \rev{setting} $u \triangleq \theta'$ and $u' \triangleq\theta$ in each transition. Thus, each bijection can be defined, as `exchanging' the auxiliary parameters with new ones in the model move,}

\begin{equation}
	g_{k \rightarrow k'}(u, \theta) = \{u', \theta'\} = \{\theta, \theta'\}\rev{.}
\end{equation}

\rev{Considering} the above, it is clear that the dimension matching constraint is satisfied and that the Jacobian is a row-wise re-ordering of the identity matrix. Therefore, the Jacobian determinant term in the acceptance ratio must be identically equal to one. The acceptance probability may now be simplified \rev{to}

\begin{equation}
	\alpha(\theta \rightarrow \theta')  = \min \Big\{ 1, \\
	\underbrace{\frac{ \pi(\theta')}{\pi(\theta)}}_{\substack{\text{Target} \\ \text{ratio}}}
	\underbrace{\frac{J(k', \theta' | k, \theta)}{J(k, \theta | k', \theta')} }_\text{Jump ratio}
	\underbrace{\frac{q(\theta | \theta')}{q(\theta' | \theta)}}_{\substack{\text{Proposal} \\ \text{ratio}}}
	\Big\}\rev{.}
	\label{eqn:AR_gen}
\end{equation}


\section{Bayesian identification of nonlinear dynamics}

\rev{With the relevant RJMCMC theory established, and the inference problem formalised, attention can be returned to the problem of sampling from the posterior density $p(\Xi, m| \mathcal{D})$. Development of this posterior over both the model terms, and their parameterisation requires the specification of both prior and proposal densities.} It is clear that there are many appropriate choices for these objects and\rev{,} in practice\rev{,} it can be expected that prior selection will be guided by the problem at hand. However, it can be shown that certain choices of the parameter proposal can lead to drastic simplification of the inference scheme.

\rev{Considering for now the case whereby the noise variance $\sigma^2$ in \eqref{eqn:sindy_bayes} is known and \rev{by} substituting the required posterior distribution into the acceptance probability \eqref{eqn:AR_gen} above, one has,}

\begin{equation}
	\alpha(\Xi \rightarrow \Xi')  = \min \Big\{ 1, \\
	\underbrace{\frac{p(\Xi', m'| \mathcal{D}, \sigma^2)}{p(\Xi, m | \mathcal{D}, \sigma^2)}}_{\substack{\text{Posterior} \\ \text{ratio}}}
	\underbrace{\frac{J(m' | m )}{J(m | m')} }_\text{Jump ratio}
	\underbrace{\frac{q(\Xi | \Xi')}{q(\Xi' | \Xi)}}_{\substack{\text{Proposal} \\ \text{ratio}}}
	\Big\}\rev{.}
	\label{eqn:the_prop}
\end{equation}

\rev{The proposal density $q$ in \eqref{eqn:the_prop} is freely chosen by the user.} A natural choice, (after Troughton and Godsill \cite{troughton1997reversible} and similar to the approach of \cite{brooks2003efficient}) is to employ the full conditional density,

\begin{equation}
	q(\Xi' | \Xi) = p(\Xi' | m', \mathcal{D}, \sigma^2)\rev{.}
\end{equation}

\rev{Substituting proposal into \eqref{eqn:the_prop} above, one finds that since,}

\begin{equation}
	p(m | \mathcal{D}, \sigma^2) = \frac{p(\Xi, m | \mathcal{D}, \sigma^2)}{p(\Xi | m, \mathcal{D}, \sigma^2)}\rev{,}
\end{equation}

the acceptance ratio can be further simplified to,

\begin{equation}
	\alpha(\Xi \rightarrow \Xi')  = \min \Big\{ 1, \\
	\underbrace{\frac{p(m' | \mathcal{D}, \sigma^2)}{p(m | \mathcal{D}, \sigma^2)}}_{\substack{\text{Ratio of model} \\ \text{posteriors}}}
	\underbrace{\frac{J(m' | m)}{J(m | m')} }_\text{Jump ratio}
	\Big\}\rev{.}
\end{equation}

This simplification is only available for particular choices of the prior $p(\Xi, m)$ that are conjugate, e.g.\ the Gaussian likelihood defined above (and defined independently to the prior over models). \rev{Moreover, it gives direct access to the marginal posteriors $p(m | \mathcal{D})$ (see \cite{pml1Book} (Chapter 6) or \cite{pml2Book} (Chapter 3) for example). Considering here, a conjugate Gaussian prior at each model order (conditioned on the model choice), one has,}

\begin{equation}
	p(\Xi | m) \triangleq \mathcal{N}(\mu_m^{(0)}, \Sigma_m^{(0)})\rev{.}
\end{equation}

\rev{Because the prior is conjugate, for a given model and noise variance, one may write,}

\begin{equation}
	\rev{p(\Xi | m,  \mathcal{D}, \sigma^2) = \mathcal{N}(\mu_m, \Sigma_m)}\rev{,}
	\label{eqn:param_post}
\end{equation}

where,

\begin{equation}
	\Sigma_m =  \left[ \frac{1}{\sigma^2} \Theta_m \tran \Theta_m + \Sigma_m^{(0) -1} \right]^{-1}\rev{,}
\end{equation}

\begin{equation}
	\mu_m = \Sigma_m \Sigma_m^{(0)-1} \mu_m^{(0)} + \frac{1}{\sigma^{2}} \Sigma_m\tran \Theta\tran \dot{x}\rev{,}
\end{equation}

are the posterior mean and covariance of the parameters for a given model $m$ (see \cite{pml1Book, pml2Book} for example). \rev{The marginal posteriors are thus available as,}

\begin{equation}
	\begin{split}
		p(m | \mathcal{D}, \sigma^2) &\propto p(m)  p(\mathcal{D} | m, \sigma^2)\rrev{,} \\
		&= p(m) (2\pi \sigma^2)^{\frac{N}{2}} |\Sigma^{(0)}_m|^{-\frac12} |\Sigma_m|^{\frac12} \exp \left( -\frac12 \left[ \sigma^{-2} \mathcal{D}\tran\mathcal{D} - \mu_m \tran \Sigma^{-1}_m \mu_m  +  [\mu^{(0)}_m]\tran [\Sigma_m^{(0)}]^{-1} \mu^{(0)}_m \right] \right)\rev{.}
	\end{split}
\end{equation}

Now assuming a zero-mean prior over the model parameters $\Xi$\footnote{Note that in the case of non-zero mean parameter priors, an additional exponentiated quadratic form will appear in the numerator and denominator of the acceptance ratio.}, the overall acceptance probability of a move from $\Xi$ to $\Xi'$ can be written,

\begin{equation}
	\alpha(\Xi \rightarrow \Xi')  = \min \Big\{ 1, \\
	\underbrace{\frac{p(m')}{p(m)}}_\text{Model prior}
	\underbrace{\frac{J(m' | m)}{J(m | m')} }_\text{Jump ratio}
	\underbrace{
	\frac{|\Sigma^{(0)}_{m'}|^{-\frac12} |\Sigma_{m'}|^{\frac12} \exp \left( \frac12 \mu_{m'}\tran \Sigma^{-1}_{m'} \mu_{m'} \right)}{|\Sigma^{(0)}_m|^{-\frac12}|\Sigma_m|^{\frac12} \exp \left( \frac12 \mu_m \tran \Sigma^{-1}_m \mu_m \right)}}_{\substack{\text{Model posterior ratio}}}
	\Big\},
	\label{eqn:bindy_ar}
\end{equation}

where several terms that do not depend on $m$ have cancelled out in the model posterior ratio. \rev{The acceptance probability in \eqref{eqn:bindy_ar}} can now be readily used to sample from the required posterior. In practice, the noise variance in \eqref{eqn:sindy_bayes} will be unknown. This problem can be simply addressed by including a Gibbs move for the noise variance given the model and the parameters, which are available following a reversible-jump move. \rev{For example an inverse-gamma prior, conjugate to the Gaussian likelihood can be employed. The Gibbs step in this case is,}

\begin{equation}
	p(\sigma^2) \triangleq \mathcal{IG}(a^{(0)}, b^{(0)})\rrev{.}
\end{equation}

Then by the conjugacy of the prior and likelihood, the posterior over the noise variance is available exactly as an inverse-gamma distribution,

\begin{equation}
	p(\sigma^2 | \Xi, m, \mathcal{D}) = \mathcal{IG}(a, b)\rev{,}
	\label{eqn:noise_gibbs}
\end{equation}

where,

\begin{equation}
	a = a^{(0)} + \frac{n}{2}\rev{,}
\end{equation}

\begin{equation}
	b = b^{(0)} + \frac{\sum_i^n (\dot{x} - \Theta(x)\Xi)^2}{2}\rev{,}
\end{equation}

where $n$ is the number of data in $\dot{x}$.

\subsection{\rev{A framework for Bayesian identification of nonlinear dynamics}}

\rev{It is useful to consider here, how the above may be used by the practitioner in order to arrive at a useful estimate of the posterior distribution $p(\Xi, m, \sigma^2 | \dot{x}, \Theta(x))$ in the context of nonlinear system identification. In practice there are several quantities that must be selected by the user. These are: the parameter and noise variance priors, the model prior, and the jump kernel. The choices of these objects will affect the cognitive biases in the inference (including the sparsity of the posterior) as well as affecting the convergence properties of the sampling scheme. Some discussion on how these objects might be selected in practice is held here.}

\rev{\textbf{Parameter and noise variance priors, $p(\Xi)$ and $p(\sigma^2)$}}

\rev{A constraint of the method introduced here is that these priors must be conditionally conjugate to the Gaussian likelihood model in \eqref{eqn:sindy_bayes}. Although this is a strong constraint it remains possible to reflect many types of prior belief with these objects. For the parameter priors, vague belief can be introduced by consideration of very wide Gaussian distributions centred around zero. Stronger belief can be imposed by the consideration of more concentrated variances. For the noise variance prior, lack of belief can be imposed by the improper parametrisation of the Inverse gamma distribution, $\mathcal{IG}(0, 0)$ (although there are potentially limitations if samples are required to be drawn from the prior). Stronger belief can be enforced by more concentrated parameterisations.}

\rev{\textbf{Model space priors, $p(m)$}}

\rev{One of the major advantages of the formulation considered here is that prior belief can be applied directly in the model space without using parameter values as a proxy. The RJMCMC scheme employed here places no restriction on the form of the prior distribution $p(m)$. Thus, very many cognitive biases can be injected into the inference scheme by the careful consideration of this parameter. Ignorance as to the model structure and level of sparsity can be trivially introduced by the consideration of a flat model prior. For example,}

\begin{equation}
	\rev{p(m)\propto 1, \quad  \forall\ m\in\mathcal{M}.}
\end{equation}

\rev{The cognitive bias of model parsimony can also be introduced. For example one could naively choose to prefer models with fewer terms (thus promoting sparsity) by the consideration of a monotonically decreasing discrete density defined over the number of terms in the model. Appropriate choices might include the geometric distribution,}

\begin{equation}
	\rev{p(m) = (1-\theta)^{d} \theta,}
\end{equation}

\rev{where $\theta$ is a hyperparameter that controls the extent to which more parsimonious models are preferred. Other types of belief could be applied by assigning probabilities to model term inclusion directly. An instructive example might be in the identification of structural dynamics in a near linear regime. One could envisage a prior structure that placed a high probability of selecting expected linear terms (corresponding to inertia, viscous damping and linear elasticity), while assigning less prior probability to nonlinear terms. The effect would be that the nonlinear terms would only appear in the posterior if there was significant evidence in the likelihood. As with all prior selection in Bayesian approaches, the choice of the model-space prior must ultimately be guided by the problem at hand and domain knowledge (or lack thereof) \cite{o2019expert}. However, the authors note that in most SINDy-type applications, practitioners expect the observed dynamics to be sparse in the library of functions. This assumption motivates the use of regularising priors in general}

\rev{\textbf{Jump kernel, $J(m'|m)$}}

\rev{The selection of the jump kernel is one of the key components of the RJMCMC and may strongly affect the convergence of the sampler. Here, we propose the following `bit-flipping' scheme. In each proposal step a random index in $[1, n]$ (where $n$ is the number of rows in $\Theta$) is selected with equal probability. If that term is present in the model at the current state of the chain then it is removed, if not then it is added. If a move is rejected then a new index is drawn, again with even probability. While this approach is computationally convenient, it is likely that for certain posterior geometries there would be significant limitations. Firstly, the proposed scheme always proposes to leave the current model. For tightly concentrated posteriors (wherein most of the probability mass is concentrated in a single model), this approach might result in low acceptance rates and inefficient exploration. A second limitation may occur in situations whereby the posterior has several modes, each corresponding to highly different model structures. In this second scenario, the proposed approach may struggle to mix from one mode to another. A potential solution would be to increase the number of indices drawn, although this could exacerbate the issues with tightly concentrated posteriors.}

\rev{An alternative to the `bit-flipping' approach employed here could be to incorporate the idea of neighbourhood proposals \cite{liang2022adaptive, caron2023structure}. These methods generalise the bit-flipping scheme above to balance exploration of the model space with exploitation near to high-probability models. These schemes typically consider moves from the current model to one in a nearby `neighbourhood' that is adaptively varied based on data or posterior probabilities. These ideas are related to the more general idea of adaptive MCMC schemes \cite{andrieu2008tutorial} that seek to improve the efficiency of MCMC samplers by including adaptive proposal mechanisms.}

\rev{\textbf{BINDy algorithm}}

\rev{The overall sampling approach is summarised in Algorithm \ref{alg:BINDy}. In the algorithm it is assumed that there is a Gaussian prior over the parameters $\Xi$ and an inverse gamma prior over the noise variance. Note that because the parameter vector does not appear in \eqref{eqn:bindy_ar}, there is no need to specify an initial condition for the parameter vector.}

\begin{algorithm}
	\caption{\rev{Bayesian identification of nonlinear dynamics}}\label{alg:BINDy}
	\begin{algorithmic}
		\Require{\rev{Parameter prior mean and variance $\mu_m^{(0)}, \Sigma_m^{(0)}$, noise variance prior parameters $a^{(0)}, b^{(0)}$, prior over models $p(m)$, jump kernel $J(m'|m)$, term library $\Theta(x)$, state derivatives $\dot{x}$, initial chain state ($m_0$, $\sigma_0^2$). }}
		\State{\rev{Initialise the chain at $m=m_0$ and $\sigma^2=\sigma_0^2$.}}
		\For{\rev{Number of samples $N$}}
		\State{\rev{Propose a model move $m \rightarrow m'$ by sampling from $J(m' | m)$.}}
		\State{\rev{Sample a new parameter vector $\Xi'$ for the model $m'$ by sampling from $p(\Xi | \mathcal{D}, \sigma^2)$ in \eqref{eqn:param_post}.}}
		\State{\rev{Accept the model and parameter move jointly with probability $\alpha(\Xi \rightarrow \Xi')$ given by \eqref{eqn:bindy_ar} else remain in place.}}
		\State{\rev{Propose a new noise variance $\sigma^2$ using the Gibbs update given by \eqref{eqn:noise_gibbs}.}}
		\EndFor
		\State{}
		\Return{\rev{Samples from $p(\Xi, m, \sigma^2| \mathcal{D})$}}
	\end{algorithmic}
\end{algorithm}

\rev{\textbf{Posterior interpretation}}

\rev{Once the practitioner has collected a sufficient set of samples (convergence of the chain can be verified by observing the trace of the parameter values in $\Xi$ for example), the posterior density can be examined in the usual Monte-Carlo fashion by counting terms and by forming histograms over model parameters. In the context of nonlinear system identification, it is often of interest to compare the performance of the identified models in \emph{simulation}. The simulation performance of a model is a better measure of the quality of the fit to data and as such represents a more robust challenge than prediction of state derivative from known states and library. Because the proposed approach here generates a probability distribution and not a single model, the simulation performance of the posterior can be evaluated sample-wise, by integrating samples from $p(\Xi, m, \sigma^2| \mathcal{D})$ forward in time (with or without added noise depending on the application). The overall simulation performance can then be established by considering the distribution over sample trajectories. Access to robust uncertainty quantification in simulation has important applications of system identification including active learning \cite{sindy-ensemble}, model predictive control \cite{lew2012robust} and structural health monitoring \cite{rogers2019towards}.}

\subsection{Related work}
\label{sec:related}

Several authors have approached the joint model structure and parameterisation problem from a Bayesian perspective. A key contribution is the work of Abdessalem et al. \cite{abdessalem2018model,abdessalem2019model} wherein an approximate Bayesian computation (ABC) method was applied to the identification of a nonlinear system from within a small number of candidate models. Although the results of the method are compelling (approximate posteriors, jointly over models and parameters), the utility of the method is limited by the extreme computational cost of ABC.

The proposed method is also not the first time that the SINDy method has been formulated in a Bayesian context. Previous attempts have made use of sparsity inducing priors \cite{fuentes2019efficient, fuentes2021equation, nayek2021spike, chen2021bayesian, hirsh2022sparsifying, more2023bayesian}. These priors are placed over parameters in $\Xi$ such that much of the posterior mass is concentrated at or around zero (thus removing them from the model with high probability). Although sparsity-inducing priors are shown to be very effective in correctly identifying model terms (by setting superfluous parameters to zero), the method cannot obtain a distributional estimate over the terms in the model independently of their parameterisation\footnote{Because sparsity inducing priors are usually defined continuously over the parameter range it is not possible to marginalise out the parameter values without setting a heuristic threshold on which parameter values should be considered insignificant.}. In contrast, the method proposed in this work differs in that prior distributions are placed over both model terms and their parameters independently. Another challenge with sparsity-inducing priors is that the inference is usually intractable and can be expensive to compute. The work in this paper is able to significantly reduce the computational burden by marginalising the parameters $\Xi$ from the acceptance ratio (after \cite{troughton1997reversible}) such that a new parameter need only be sampled when updating the noise variance.

A related work is the recent contribution of Fung et al. \cite{fung2024rapid}. The authors propose a Bayesian formulation of the SINDy method by tracing uncertainty through the nonlinear library and the differential operator in order to account for noise on observed $x$. The authors then make Gaussian approximations to recover the model evidence in a Bayesian manner. The authors then propose to greedily select terms that maximally increase the model evidence until no more such terms are available. This results in a single model structure estimate with a posterior distribution over its parameterisation. The method proposed in this differs in that a joint posterior distribution over both model terms and their parameterisation is available explicitly, thanks to the reversible-jump sampler. The proposed approach is also not limited to greedily maximising the model evidence, which can be important in situations whereby the posterior is multimodal in model space.

The method presented herein is also not the first use of RJMCMC in the Bayesian identification of dynamic models. Troughton and Godsill \cite{troughton1997reversible} and Dahlin et al.\  \cite{dahlin2011robust} consider an RJMCMC scheme for the identification of linear autoregressive models, and a RJMCMC scheme is employed in \cite{tiboaca2014bayesian} to select between two candidate nonlinear models. \rev{Recently in \cite{cox2023sparse}, an RJMCMC approach is used to select terms in a state-space formulation using Kalman filtering techniques. The paper has the advantage that partial state observation is handled naturally however the technique is limited to linear systems only.  However, to the author's best knowledge, the present work represents the first application of RJMCMC to the SINDy-type formulation of \cite{sindy}, and the first to explicitly enumerate a full posterior distribution over both model terms and their parameterisation jointly in that context.}

Another related contribution is the uncertainty quantification enabled by the ensemble-SINDy (E-SINDy) method of Fasel et al. \cite{sindy-ensemble}. E-SINDy enables heuristic UQ through bootstrapping and (robust) averaging (b(r)agging) on subsets of model terms and data. In this way, the authors are able to give heuristic estimates of quantities such as term inclusion probabilities and bootstrapped distributional estimates of model parameters. While these quantities are undoubtably useful to practitioners, their heuristic nature makes their theoretical interpretation difficult. Nevertheless, E-SINDy is a natural benchmark comparison for the method proposed in this work. In all forthcoming case studies, both methods are demonstrated.

\section{Results}

In order to demonstrate the effectiveness of the proposed method, three benchmark case-studies are presented here. Results from both the proposed method and E-SINDy \cite{sindy-ensemble} are presented and compared in terms of their confidence in identifying the true dynamics.

\rev{The prior structure of the inference is set up initially to be weakly informative.} In particular, the values,

\begin{equation}
	\mu_m \triangleq \B{0}^{(0)}, \quad \Sigma_m^{(0)} \triangleq 10^3 \times I\rev{,}
\end{equation}

\begin{equation}
	a^{(0)} \triangleq 0, \quad b^{(0)} \triangleq 0\rev{,}
\end{equation}

are chosen. \rev{The prior distribution over models is selected initially to be flat with $p(m)\propto 1, \  \forall\ m\in\mathcal{M}$. The authors would note that this choice does not promote model parsimony a priori}. The chain is initialised with all terms in $\Theta$ present in the model and with the noise variance set to $\sigma^2 = 1$. The Gibbs sampler is run for $6\times10^3$ steps in total, with the first thousand samples discarded to remove the effects of any transient `burn-in' behavior. \rev{The authors remark that this length of chain appears more than adequate in terms of convergence to the target distribution (See the appendices for a numerical study).}

The E-SINDy results presented hereafter are generated using the publicly available `pysindy' library \cite{de2020pysindy}. In all cases, $5\times10^3$ models are sampled (in line with the length of the chains in the RJMCMC) and both data and library ensembling are activated, with the number of candidates dropped in each sample set to 1. In both cases, the sequentially-thresholded least-squares (STLSQ) algorithm is used to perform the sparse regression. For more detail on the E-SINDy method and for interpretation of these hyperparameters the interested reader is directed to the original article \cite{sindy-ensemble} and the `pysindy' documentation \cite{de2020pysindy}.

\subsection{Legendre polynomials}

\begin{figure}
	\captionsetup{labelfont={color=\revcolorflag}}
	\centering
	\includegraphics[width=\fw]{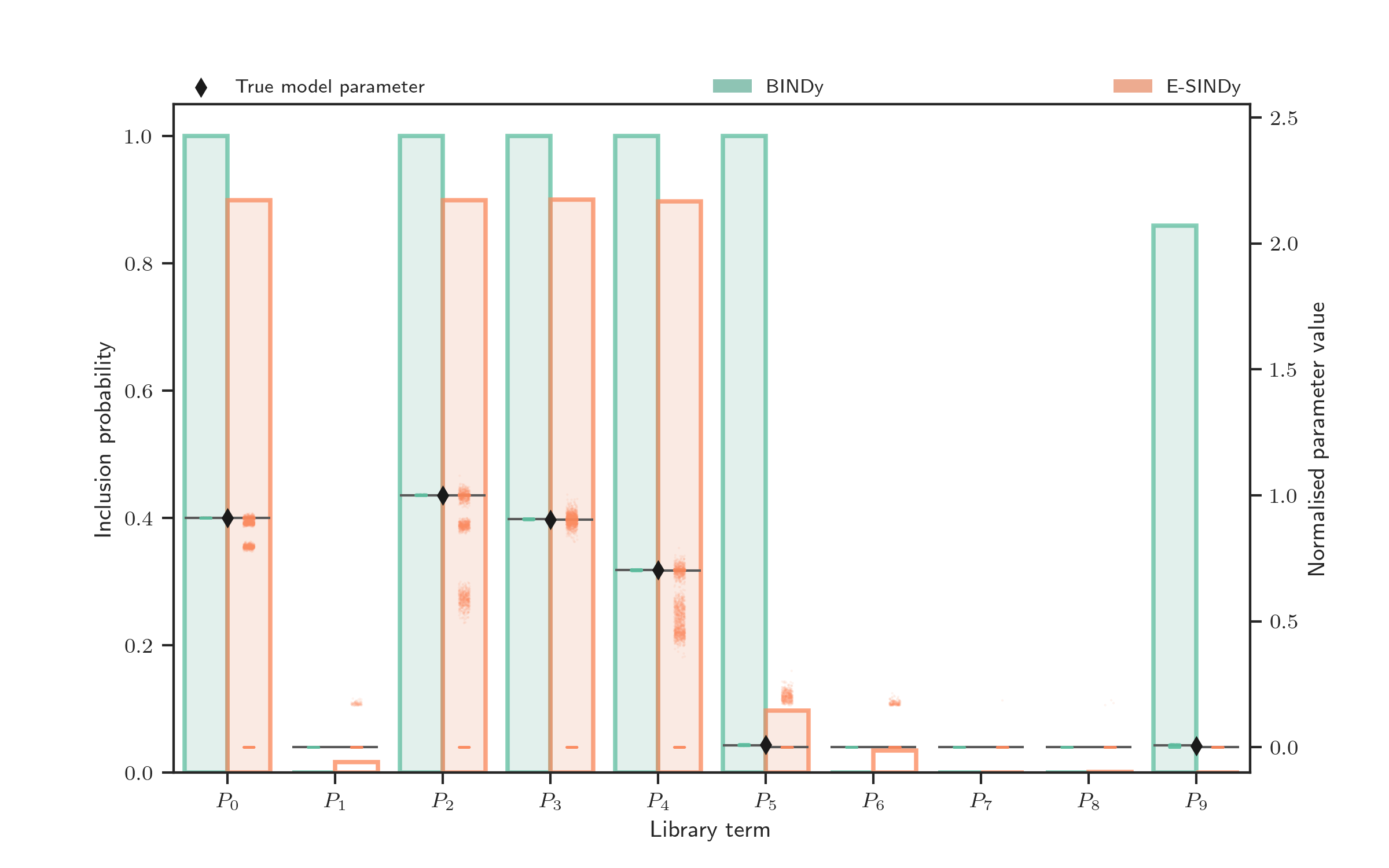}
	\caption{Comparison between RJMCMC and SINDy in quantifying model uncertainty for the toy polynomial data. In the plot, the bar-charts depict model inclusion probability (left axis), point plots depict samples of model parameters (right, log axis). Horizontal bars depict median parameter values, black diamonds depict true values of parameters in the underlying data-generating model.}
	\label{fig:poly_post}
\end{figure}

\begin{figure}
	\centering
	\includegraphics[width=\fw]{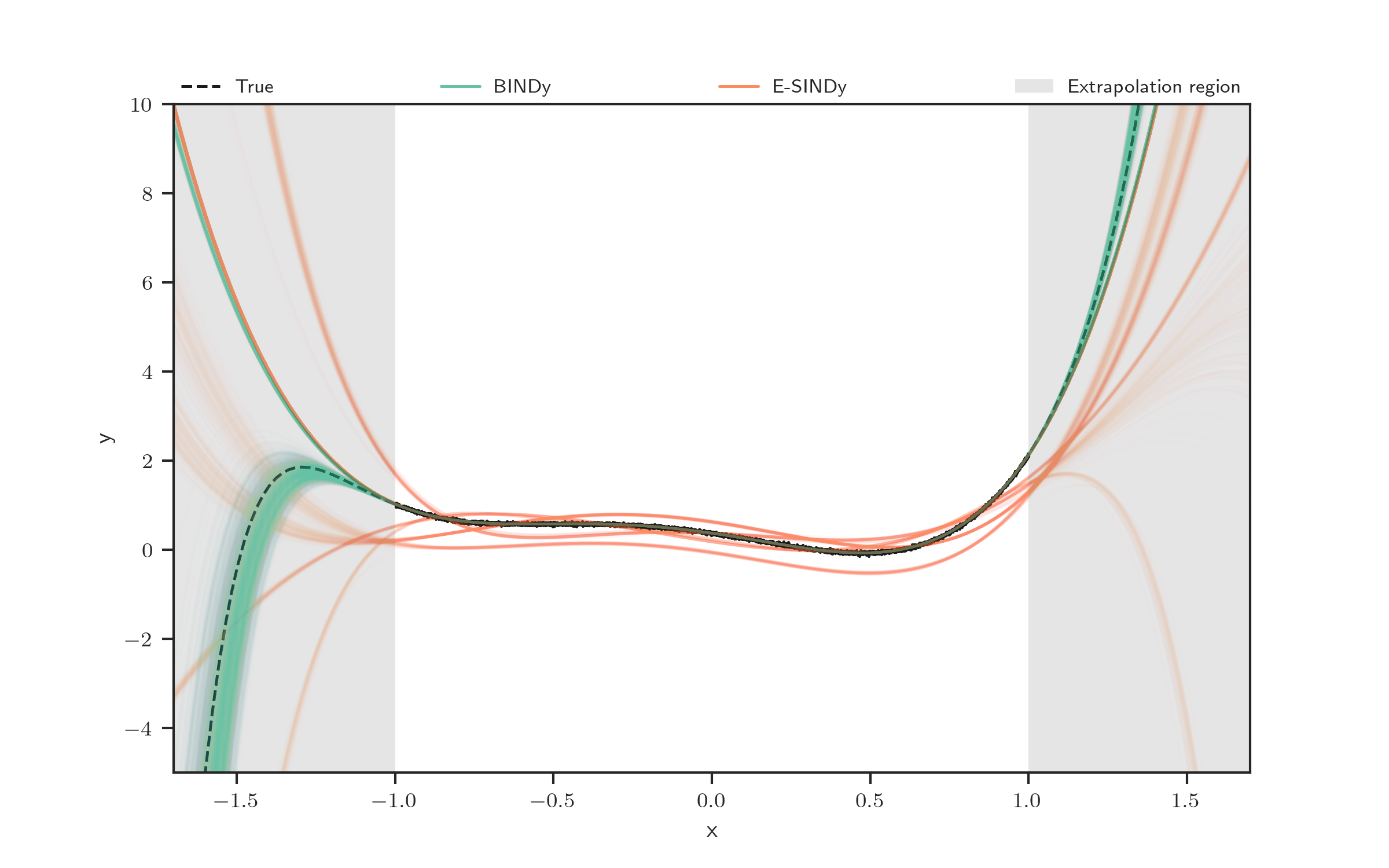}
	\caption{Samples from the identified posterior distribution (BINDy, green) and ensemble (E-SINDy, orange) in both interpolation and extrapolation regimes.}
	\label{fig:poly_traj}
\end{figure}

The first case-study is a static sparse polynomial regression problem in which $\Theta(x)$ is static and is set to be the first 10 Legendre polynomials on the interval $x \in [-1, 1]$. In order to demonstrate the recovery of small-valued parameters, a coefficient vector is selected with both zero and near-zero elements. \rev{The coefficient vector was generated randomly by the following scheme:}

\begin{itemize}
	\item \rev{10 random parameters were sampled uniformly on [0,1].}
	\item \rev{4 randomly selected parameters (with replacement) were set to 0.}
	\item \rev{2 randomly selected parameters (with replacement) were multiplied by 0.01 to drastically reduce their magnitude.}
	\item \rev{All parameters are rounded to 3 decimal places (so that they can be exactly reproduced in this document).}
\end{itemize}

\rev{The resulting coefficient vector is given by,\footnote{\rev{For clarity and readability, the results of the case study here considers only a single parameterisation of this polynomial. An additional numerical study considering many such parameterisations is presented in the appendices.}}}

\begin{equation}
	\Xi^* \triangleq \left[0.549, 0, 0.603, 0.545, 0.424, 0.006, 0, 0, 0, 0.004\right]\tran\rev{.}
	\label{eqn:polycoeff}
\end{equation}

The target of the regression is then generated as,

\begin{equation}
	y \triangleq \Theta(x)\Xi^* + \epsilon \quad \epsilon \sim \mathcal{N}(\bm{0}, \sigma^2_n I)\rev{,}
\end{equation}

whereby Gaussian noise is added to the target data with standard deviation equal to 5\% of the root-mean-square (RMS) of the nose-free data.

The performance of both E-SINDy and the proposed method are compared in Figure \ref{fig:poly_post}. As can be seen in the figure, Both the proposed method and E-SINDy methods are able to correctly assign low inclusion probability to the terms in $\Xi^*$ that are equal to zero. \rev{The posterior samples generated by BINDy are tightly concentrated around the true values such that they are virtually indistinguishable from their median.} However, the STLSQ algorithm is unable to identify terms with parameters lower than the threshold parameter $\lambda$, which in this case study is set to the pysindy default value of 0.1. While this is a simplified example, it serves to demonstrate a particular pathology, which may appear when using parameter value as a proxy for term importance. However, the E-SINDy method shows multi-modality, producing some samples in clusters away from the true value. This can be attributed to the library bagging operation. Every subset of the library functions produces a sample (none are rejected - there is no Metropolisation step). This has the effect that some samples are generated with important terms missing, biasing the parameter values of the remaining terms.

\subsection{Lynx-hare population dynamics}

\begin{figure}
	\captionsetup{labelfont={color=\revcolorflag}}
	\centering
	\includegraphics[width=\fw]{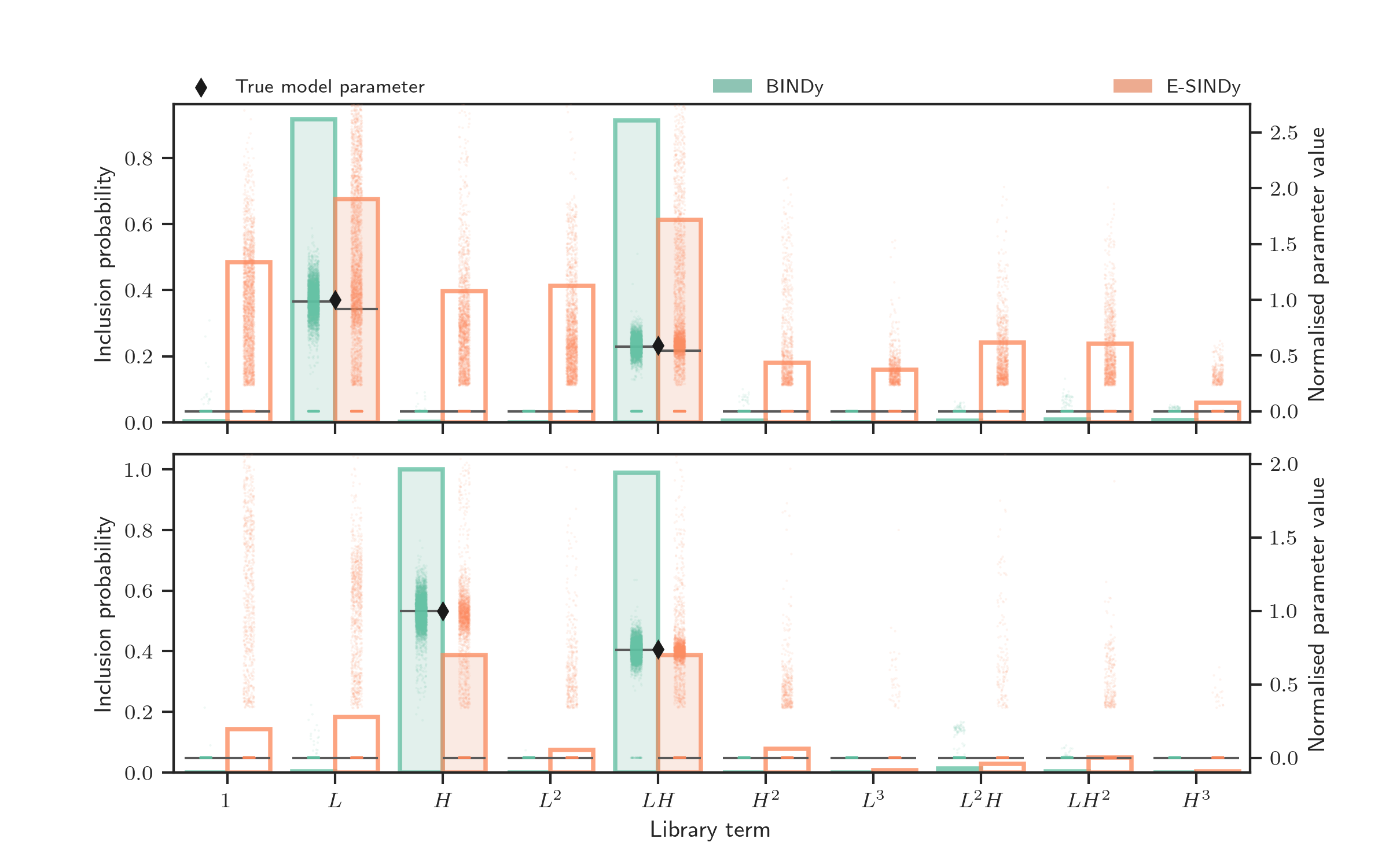}
	\caption{Comparison between RJMCMC and SINDy in quantifying model uncertainty for the lynx-hare population example. In the plot, the bar-charts depict model inclusion probability (left axis), point plots depict samples of model parameters (right, log axis). Horizontal bars depict median parameter values, black diamonds depict maximum likelihood values of parameters in the Lokta-Volterra dynamics.}
	\label{fig:lv_post}
\end{figure}

\begin{figure}
	\centering
	\includegraphics[width=\fw]{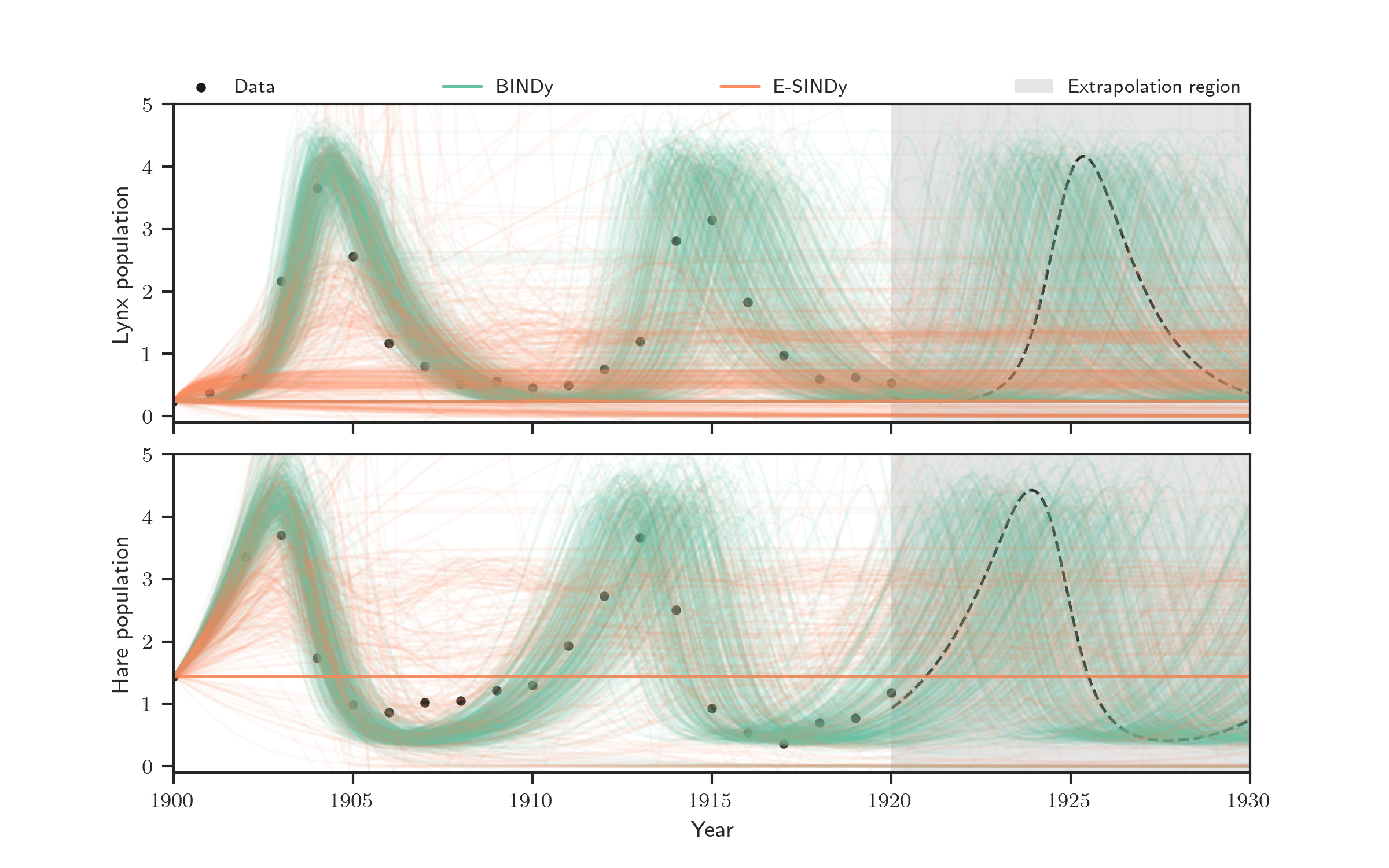}
	\caption{Samples from the identified posterior distribution (BINDy, green) and ensemble (E-SINDy, orange) in both interpolation and extrapolation regimes for the lynx-hare population example.}
	\label{fig:lv_traj}
\end{figure}

The second case-study considers the well-studied lynx-hare population dynamics dataset. This dataset consists of the number of pelts collected for hares and lynxes annually between 1900 and 1920 by the Hudson Bay Company and is considered to reflect the population level of the two species \cite{sindy-ensemble}. As such, the dataset is often used as a canonical example of data generated by the Lokta-Volterra predator-prey population model. Let $L$ and $H$ be the number of Lynx and Hare in the population respectively, then their population dynamics are governed by the first order ordinary differential equation,

\rev{
	\begin{equation}
		\begin{split}
			\frac{dL}{dt} &= c_{11} L + c_{12} H L \\
			\frac{dH}{dt} &= c_{21} H + c_{22} H L
		\end{split}\rev{,}
	\end{equation}
}

\rev{where the $c_{ij}$ are unknown constants in the model.} The dataset represents a significant challenge in that only 21 measurements are available.

Here, the derivatives of the lynx and hare populations are computed numerically in accordance with the approach taken in \cite{sindy-ensemble}, the columns of $\Theta$ are also normalised and a threshold value of 0.19 for $\lambda$ is used. For $\Theta(x)$, library of polynomial terms, including interactions up to third order is used. The authors note that this constitutes a more challenging identification task\footnote{The sizes of the respective model spaces increase from $2^{6}$ to $2^{10}$ when considering terms up to third order.} than the one presented in \cite{sindy-ensemble}, where only terms up to second order are considered.

The distributions identified by the two methods on the lynx-hare dataset are depicted in Figure \ref{fig:lv_post}. In the figure, it can be seen that both models are assigning high probability to terms in the Lokta-Volterra model. However, the probability mass is much more concentrated around the these terms in the BINDy posterior, compared to those in the E-SINDy ensemble. Once again, it can be seen that the parameter values are tightly distributed around the maximum likelihood values for both methods (assuming Lokta-Volterra dynamics), although the E-SINDy method produces a number of samples close to the cutoff threshold. Particularly concerning, is that the median parameter value for all terms in the hare evolution (as identified by ensemble SINDy) is zero. This corresponds to the STLSQ algorithm in E-SINDy removing all parameters and returning a zero model much of the time. These constant dynamics (with no terms in the model) identified by the SINDy ensemble are evident in Figure \ref{fig:lv_traj} which plots time histories (integrated forward in time numerically, from known initial conditions) of samples from both BINDy and E-SINDy. By comparison, the samples from the BINDy posterior represent the dynamics well, given the scarcity of data available. In extrapolation, neither model is able to closely follow the mean dynamics (computed as the least squares estimate of $a,b,c,d$, assuming the model structure to be known). However, the uncertainty in the dynamics offered by the BINDy posterior is visually far more compelling, expanding in both the location and magnitude of the population peaks for each species--as might be expected. Access to meaningful uncertainty quantification for real datasets in the low-data regime is of huge importance, for example when modelling epidemiological dynamics, wherein predicting the height and timing of peaks is critical.

\subsection{Lorenz attractor dynamics}

\begin{figure}
	\captionsetup{labelfont={color=\revcolorflag}}
	\centering
	\includegraphics[width=\fw]{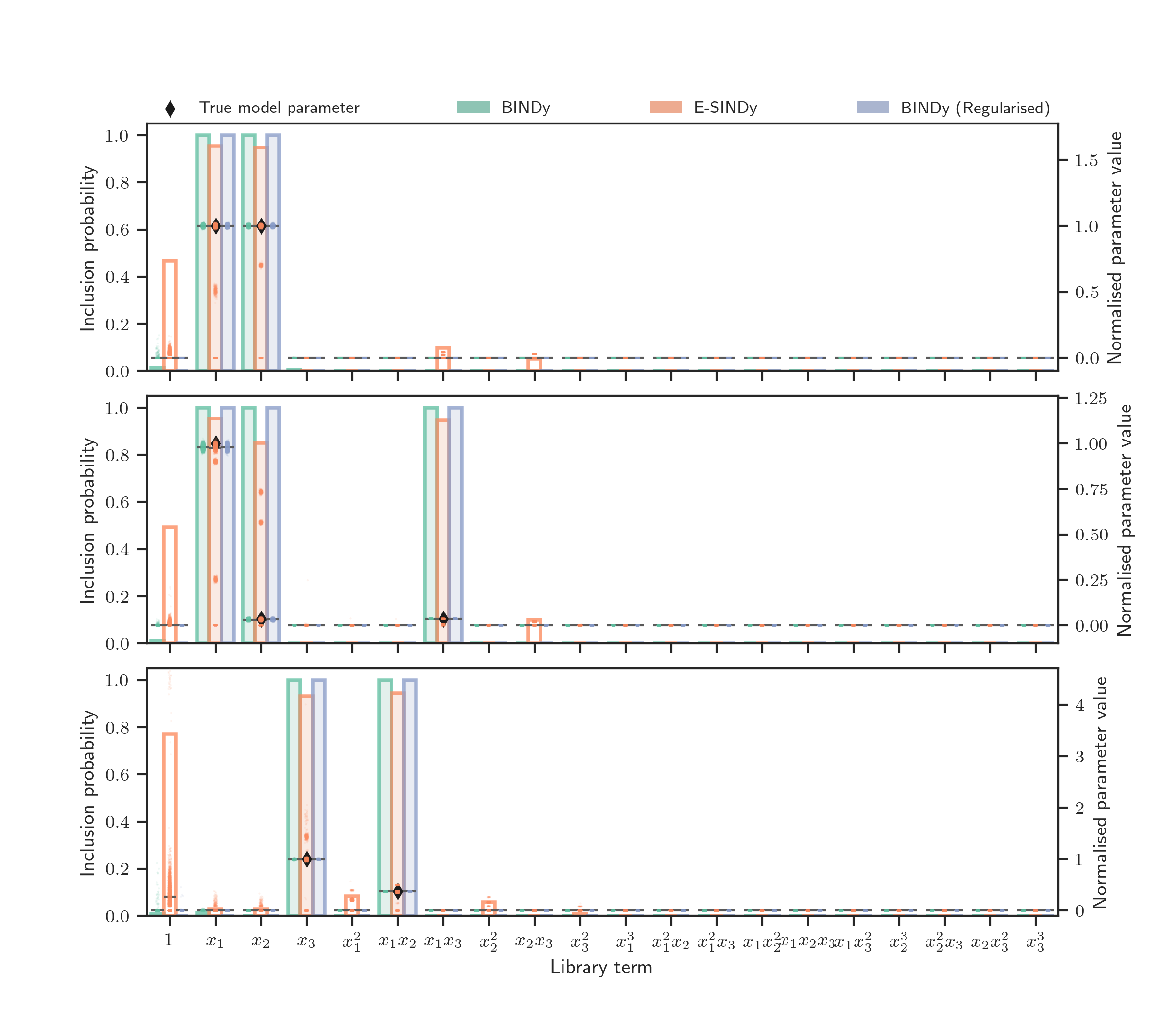}
	\caption{Comparison between RJMCMC and SINDy in quantifying model uncertainty for the Lorenz attractor example. In the plot, the bar-charts depict model inclusion probability (left axis), point plots depict samples of model parameters (right, log axis). Horizontal bars depict median parameter values, black diamonds depict true values of parameters in the underlying data-generating model.}
	\label{fig:lorz_post}
\end{figure}

\begin{figure}
	\centering
	\includegraphics[width=\fw]{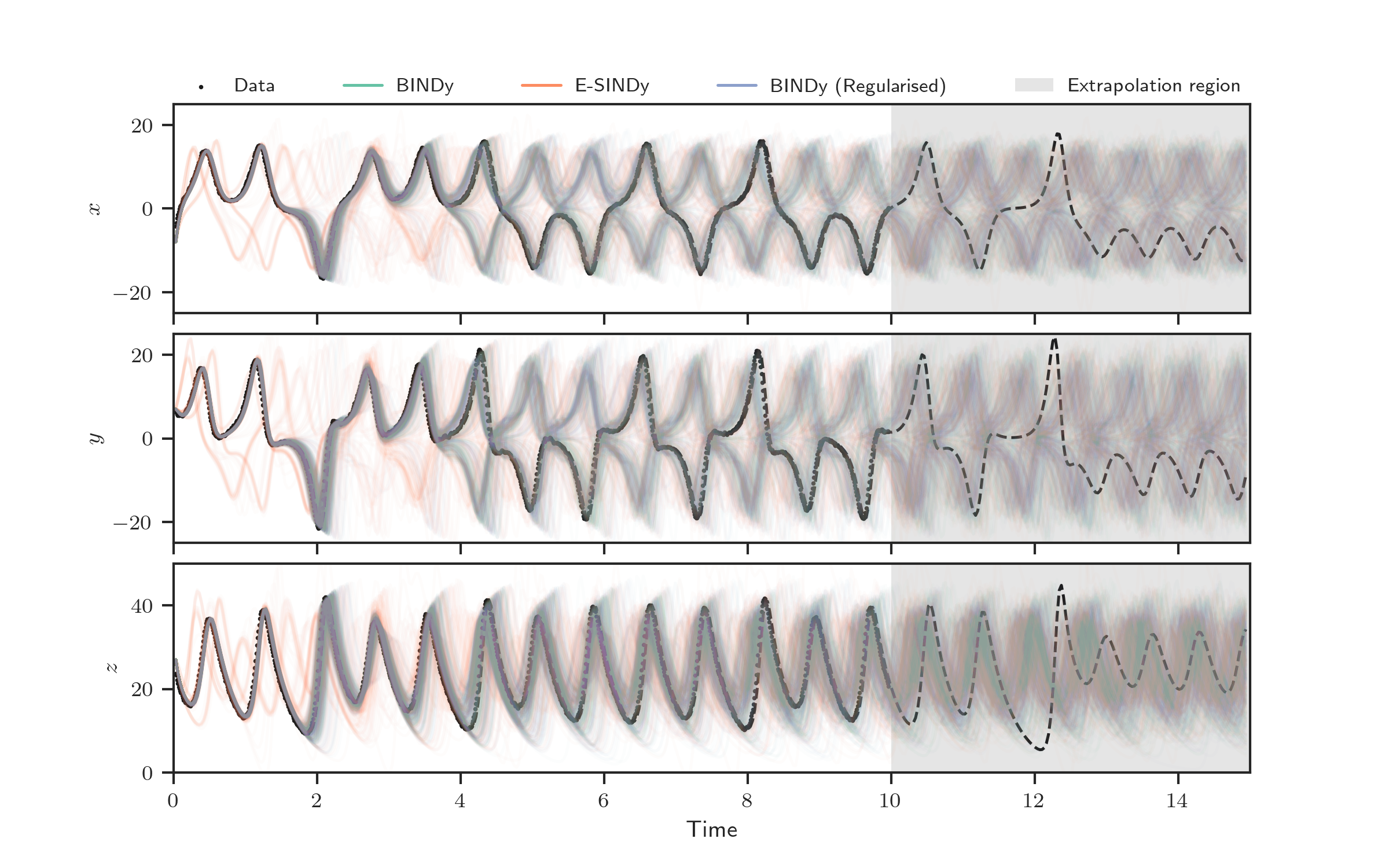}
	\caption{Samples from the identified posterior distribution (BINDy, green) and ensemble (E-SINDy, orange) in both interpolation and extrapolation regimes for the Lorenz attractor example.}
	\label{fig:lorz_traj}
\end{figure}


The third case study considers the chaotic dynamics of the Lorenz attractor. The dynamics of the Lorenz attractor are well studied in the SINDy literature \cite{sindy,sindy-ensemble}, and have become a benchmark system for advancements to the SINDy method. The dynamics are driven by the following system of ordinary differential equations\rrev{:}

\begin{equation}
	\begin{split}
		\frac{dx}{dt} &= 10(x_2-x_1)\rrev{,} \\
		\frac{dy}{dt} &= x_1(28-x_3)\rrev{,} \\
		\frac{dz}{dt} &= x_1x_2 - \frac83 x_3\rrev{.}
	\end{split}
\end{equation}

For the case study in this work, the Lorenz equations above are simulated for 10 seconds (with a further 5 seconds of unseen data for extrapolation) from an initial condition of $[x_1,x_2,x_3]\tran = [-8,7,27]\tran$ at a sample rate of $10^2$ Hz. Overall, $10^3$ points are available for training. As before, $\Theta(x)$ is set as a polynomial library, considering terms up to third order. Before training, noise at 2.5\% RMS is added to the state variables. The state derivatives are computed numerically using a polynomial smoothed finite-difference scheme with a difference order of 2, a window size of 5 and a polynomial order of 3. \rev{In order to reduce the error introduced by the numerical computation of the derivatives close to the edges of the signal, a few time points\footnote{\rev{Dropping $2n+1$ points from each end for an $n$th-order finite difference scheme is the default used in pysindy \cite{de2020pysindy} and here. After \cite{sindy-ensemble}, a 2nd order finite difference scheme is used.}} are dropped from the signal endpoints.} For the SINDy results, the threshold parameter is set to 0.2 (as is done in \cite{sindy-ensemble}). All other parameters are unchanged from the previous case-studies.

\rev{For this third case study, the effect of the prior over models $p(m)$ is also considered. In addition to the weakly informative prior used in the previous case-study examples, a regularising geometric prior over the total number of non-zero elements in the model is also considered. The probability mass function of the regularising prior is given by,}

\begin{equation}
	p(m) = (1-\theta)^{d} \theta\rev{,}
\end{equation}

where $d$ is the number of terms in model $m$, and $\theta$ is a hyperparameter, set to the strongly regularising value of 0.99 in this work. The effect of this prior over models is to strongly penalise (by assigning low prior probability to) models with many terms.


The result of the UQ for the Lorenz equations is compared in Figure \ref{fig:lorz_post} for BINDy (with and without the regularising prior) and E-SINDy. It can be seen in the figure that all three approaches have been highly successful in assigning high probability to the terms in the data-generating model. As before, sampled model trajectories (integrated forward in time) are plotted for all three methods in Figure \ref{fig:lorz_traj}. The chaotic nature of the Lorenz equations makes tracking the true dynamics highly challenging. The well-known `lobe-switching' behavior of the dynamics means that small errors can quickly cause significant deviations as the dynamics become governed by a different local attractor. This effect is clearly present in the identified distributions, which quickly become multimodal after a few oscillations. Qualitatively, it is observed that both BINDy and regularised BINDy samples remain in step with the true dynamics for the longest, with E-SINDy the first to diverge. This divergence can be seen \rev{almost immediately} where some orange lines deviate quickly from the observed data. Indeed, this is supported by the numerical results in Table \ref{tab:lorz_traj} which show lower error statistics for the BINDy and regularised BINDy approach. In the extrapolation region, the distributions have become very diffuse, in keeping with what might be expected for uncertain forecasting of a chaotic attractor such as the Lorenz equations.

\begin{table}
	\centering
	\caption{Posterior results for the Lorenz attractor case-study. Term inclusion probabilities $p(\Theta_i(x)|\mathcal{D})$ and expected parameter values $E[\Xi_i | \mathcal{D}]$ (and standard deviations) for each of the three considered methods.}
	\label{tab:lorz_post}
	\begin{tabular}{ll|ll|ll|ll}
		\multirow{2}{*}{Equation}               & \multirow{2}{*}{Term} & \multicolumn{2}{c}{BINDy} & \multicolumn{2}{c}{E-SINDy} & \multicolumn{2}{c}{BINDy (regularised)}                                                              \\
		                                        &                       & $p(\Theta_i|\cdot)$       & mean (std)                  & $p(\Theta_i|\cdot)$                     & mean (std)       & $p(\Theta_i|\cdot)$ & mean (std)        \\\hline\hline
		\multirow{2}{*}{$10(x_2-x_1)$}          & $x_1$                 & 1.00                      & -10      (0.0759)           & 0.95                                    & -8.74    (3.88)  & 1.00                & -10      (0.0742) \\
		                                        & $x_2$                 & 1.00                      & 10       (0.0655)           & 0.95                                    & 9.33     (2.29)  & 1.00                & 10       (0.0641) \\\hline
		\multirow{3}{*}{$x_1(28-x_3)$}          & $x_1$                 & 1.00                      & 27.4     (0.314)            & 0.95                                    & 23.9     (9.43)  & 1.00                & 27.4     (0.317)  \\
		                                        & $x_2$                 & 1.00                      & -0.872   (0.109)            & 0.85                                    & 1.20      (5.82) & 1.00                & -0.875   (0.109)  \\
		                                        & $x_1 x_3$             & 1.00                      & -0.985   (0.0074)           & 0.94                                    & -0.889   (0.267) & 1.00                & -0.985   (0.0074) \\\hline
		\multirow{2}{*}{$x_1x_2 - \frac83 x_3$} & $x_3$                 & 1.00                      & -2.66    (0.0169)           & 0.93                                    & -2.46    (0.973) & 1.00                & -2.66    (0.0167) \\
		                                        & $x_1 x_2$             & 1.00                      & 0.995    (0.0038)           & 0.94                                    & 0.946    (0.239) & 1.00                & 0.995    (0.0039) \\
	\end{tabular}
\end{table}

\begin{table}
	\centering
	\caption{Median, Mean and standard deviations of sample mean-squared errors computed from the trajectories in Figure \ref{fig:lorz_post}.}
	\label{tab:lorz_traj}
	\begin{tabular}{l|lll|lll|lll}
		\multirow{2}{*}{Equation} & \multicolumn{3}{c}{BINDy} & \multicolumn{3}{c}{E-SINDy} & \multicolumn{3}{c}{BINDy (regularised)}                                               \\
		                          & median                    & mean                        & std                                     & median & mean & std  & median & mean & std  \\\hline\hline
		$10(x_2-x_1)$             & 73.7                      & 73.5                        & 18.1                                    & 82.8   & 83.8 & 19   & 74.8   & 75.2 & 14.4 \\
		$x_1(28-x_3)$             & 100                       & 98.1                        & 22.9                                    & 110    & 112  & 25   & 100    & 101  & 18.2 \\
		$x_1x_2 - \frac83 x_3$    & 56.4                      & 62.2                        & 18.5                                    & 79.3   & 86.5 & 36.8 & 59     & 65   & 20   \\
	\end{tabular}
\end{table}

\section{Discussion}

In all three case studies, strongly parsimonious models were identified by the proposed methodology. Counterintuitively, these models were assigned high posterior probability despite a flat prior over the model space. It is interesting to consider how this can be the case, for which the authors offer two explanations. On examination of the acceptance ratio in \eqref{eqn:bindy_ar}, one finds that the ratio is proportional to $|\Sigma^{(0)}_{m'}|^{-\frac12} / |\Sigma^{(0)}_{m}|^{-\frac12}$. Intuitively, this ratio has a regularising effect that penalises larger models that have the same fit to the data due to the differing dimensions of the prior covariance. This effect scales with the parameter covariance meaning that wide priors (like those considered in the case studies here) have a stronger regularising effect. Another important consideration is the correlations between the columns of $\Theta(x)$ and the state derivatives $\dot{x}$. These correlations depend on the data, the system under study and chosen library of basis functions. Many, highly correlated terms can complicate the identification procedure with many models able to produce predictions close to the observed data. In contrast, in the case where only a few terms have high correlation, the identification of models with few terms is simplified. This effect could also explain the parsimonious models in the first two case studies.

As with any Bayesian approach, the specification of prior distributions plays an important role. A limitation of the current approach is that the priors over the parameters themselves are required to be conjugate. For most applications this is not expected to be an issue, although there are several situations (for example bounded parameter spaces or monotonic functions) where this requirement could become restrictive. The authors cannot presently envisage a solution to this issue that does not come at considerable computational cost, should alternative, non-conjugate priors be desired.

The choice of prior over models $p(m)$ may be critical in enforcing parsimony. Here, a strongly regularising geometric distribution parameterised by the number of terms in the model has been applied successfully. However, vastly many more parametrisations are possible. It is expected that prior selection in practice will be driven by expert domain knowledge, wherever possible. For example, one could set the prior over each term directly, assigning high prior probability to terms that are expected to be in the model and low probabilities otherwise. Alternatively one could promote parsimony in some types of library functions while permitting dense models in other types. The flexible nature of the Bayesian formulation presented in this work, allows modellers to encode both domain knowledge and cognitive biases \emph{a priori}. \rev{An interesting avenue for further investigation is hierarchical prior structures whereby the model-space prior is permitted to be conditional on the values its parametrisation.}

One could even imagine a library tempering scheme (similar in some ways to the sequential library and parameter bagging in \cite{sindy-ensemble}) whereby model terms with low probability are tempered out of the prior during sampling. Practically this would enable more efficient sampling (as more model moves are likely to be accepted) as well as permit the operator to start the sampler with a vastly greater (or even infinite) library of terms, with the expectation that all but those in the typical set would be tempered out. Such a scheme would bear some resemblance to well-studied sequential Monte-Carlo (SMC) samplers \cite{chopin2020introduction}, for which a great deal of analysis has already been conducted. However, the application of SMC samplers to domains whereby parameters are permitted to change (such as in the interacting scheme of Jasra \emph{et al.} \cite{jasra2008interacting}) has not been as widely studied and remains an open challenge.

A related area for future investigation is the choice of the `jump' kernel $J(m' | m)$. The `bit-flipping' method applied herein has proved to be effective in the case-studies considered so far, but it might be ineffective when significant multi-modality is expected in the posterior, for example when very differently parameterised models describe the data well, low probability  intermediate steps might inhibit exploration of the posterior. Instead, it is interesting to imagine more sophisticated schemes that could better suit problems of this type. Also interesting are so-called `non-reversible' jump methods \cite{gagnon2020nonreversible} that use an adaptive scheme to ensure that the posterior is explored efficiently. \rev{Another possibility for the jump kernel is to include an approach based on the adaptive neighbourhood proposals of \cite{liang2022adaptive,caron2023structure}. For very large model libraries (whereby evaluation of a model with all the terms included may be computationally infeasible), such approaches may give practitioners a method to retain a computationally viable sampler, by considering model moves only to `nearby' models.}

\rev{The computational complexity of the proposed approach is of interest. The core computational cost in each iteration of the proposed algorithm is the matrix inversion required to perform the conjugate update of the proposed parameters in each iteration of the sampler. Because this cost is (naively) cubic $O(d^3 + N d^2)$ for an RJMCMC move to a model order of $d$ (from a model order less than $d$) given $N$ observations, the worst-case analysis is that the method is $O(D^3)$ for a library of size $D$ terms. However, because we expect many of the samples in the posterior to be sparse $(d<<D)$ the actual computational complexity (per sample) is likely to be significantly lower than this. Despite the lack of a concrete computational complexity, the authors remark that empirically the method is extremely fast. The authors are able to run the sampler at over 200 iterations per second on a single core of a Dell XPS laptop with an i7 processor.}

\rev{Although the case studies in this paper have focussed on systems of first order ODEs, it is of interest to consider applications to other types of dynamical systems. Since their initial conception, SINDy-type methods have been extended to include PDEs \cite{sindy-partial}, implicit dynamics \cite{sindy-pi}, discrete dynamics and weak PDE solutions \cite{sindy-weak}. Because the approach here can be thought of as a replacement for the sparse regression at the core SINDy modelling there is no reason that it cannot be applied in any of these cases. However, there are some important considerations. For example in the case on implicit dynamics, care would need to be taken to treat the ill conditioning of the implicit dynamics in a robust probabilistic manner. The authors see these areas as very interesting areas for future investigation. It is also interesting to imagine the application of the proposed method to systems of very many differential equations. Such high-dimensional problems pose a challenge in that the libraries under consideration would grow combinatorially with the number of degrees of freedom. In this setting, the adaptive neighbourhood proposals in \cite{liang2022adaptive,caron2023structure} may encourage better mixing in the chain.}

A key assumption in SINDy methods is that the dynamics are described as $\dot{x} \approx \Theta(x)\Xi$. Implicit in the least-squares formulation of \cite{sindy} (as well as almost every other approach for learning sparse $\Xi$) is the Gaussian noise model in \eqref{eqn:sindy_bayes}. The assumption of Gaussian noise on $\dot{x}$ only, with $x$ (or at least $\Theta(x)$) noise free, is unlikely to hold for many practical applications -- especially in the low-data, high-noise regime. The calculation of $\dot{x}$ numerically from noisy $x$ is likely to only compound this issue, particularly for second-order dynamical systems where the numerical derivative must be computed twice. It is certainly possible to envisage more advanced statistical descriptions of the noise model, whereby noise enters the observed $x$ is transformed nonlinearly by $\Theta$ and by the numerical differentiation in computing $\dot{x}$ (such an attempt is made in \cite{fung2024rapid}). However, such a formulation is likely to make the inference described in this work impractically expensive. Despite the strong assumption of the noise model in \ref{eqn:sindy_bayes}, the authors are encouraged by the strong performance of the proposed method in both numerical and experimental case-studies. One potential explanation for this performance is that the noise variance parameter $\sigma^2$ is freely able to inflate during the sampling. This can have the effect of `swallowing' additional variance that may be present in a more accurate likelihood model. The result is a Gaussian approximation to the likelihood function (in a manner similar to \cite{fung2024rapid}).

\section{Conclusion}

In this work, a novel method for Bayesian identification of nonlinear dynamic systems in a SINDy-like manner has been proposed, considering a joint inference over both model terms and their parameterisation. It is the argument of this paper that such a posterior over models is more useful than single (or even an ensemble of) sparse models as it permits the operator to make probability-informed choices, and propagate meaningful uncertainty to further analyses.

Overall the proposed methodology has several advantages in heuristic quantification of uncertainty over ensemble-based SINDy, including prior choice over models and a valid distributional formulation. The Bayesian formulation has been shown to be effective in three benchmark case studies, correctly assigning high posterior probability to terms in the underlying differential equations.


\section*{Authors' contributions}

MDC, Conceptualisation (Equal), Methodology (Equal), Software (Lead), Data Curation (Lead), Writing - Original Draft (Lead), Writing - Review \& Editing (Supporting)

TJR, Conceptualisation (Equal), Methodology (Equal), Software (Supporting), Data Curation (Supporting), Writing - Original Draft (Supporting), Writing - Review \& Editing (Lead), Supervision (Lead), Funding acquisition (Lead)

\section*{Conflict of interest declaration}
The authors declare there are no competing interests in this work.

\section*{Funding}

The authors of this paper gratefully acknowledge the support of the UK Engineering and Physical Sciences Research Council (EPSRC) via grant references EP/W005816/1 and EP/W002140/1. For the purpose of open access, the authors have applied a Creative Commons Attribution (CC BY) licence to any Author Accepted Manuscript version arising.

\bibliographystyle{unsrtnat_keith}
\bibliography{BINDy.bib}

\newpage
\appendix
\section{\rev{Effect of parameterisation in identification of small valued-parameters}}

\rev{In the first case study example, a single parameterisation of the Legendre polynomial was considered. Here, the effect of this parameterisation is studied. \rrev{Both BINDy and E-SINDy are applied to 100 parameterisations of the Legendre polynomial in \eqref{eqn:polycoeff}.} Parameters are sampled uniformly in the range [0,1] and rounded to three decimal places (as was done in the generation of the coefficients in \eqref{eqn:polycoeff}). Of the sampled parameters, 4 were set to have zero magnitude and a further two were set to have a small magnitude by multiplication by 0.01. In order to aid in the interpretation of the results, the indices of the zero-valued and small-valued parameters were fixed to be the same as those of in the first case study ($[P_1, P_6, P_7, P_8]$ and $[P_5, P_9]$ respectively). Added Gaussian noise was also re-sampled at a level of 5\% RMS in every case.}

\begin{figure}[ht]
	\captionsetup{labelfont={color=\revcolorflag}}
	\centering
	\includegraphics[width=\fw]{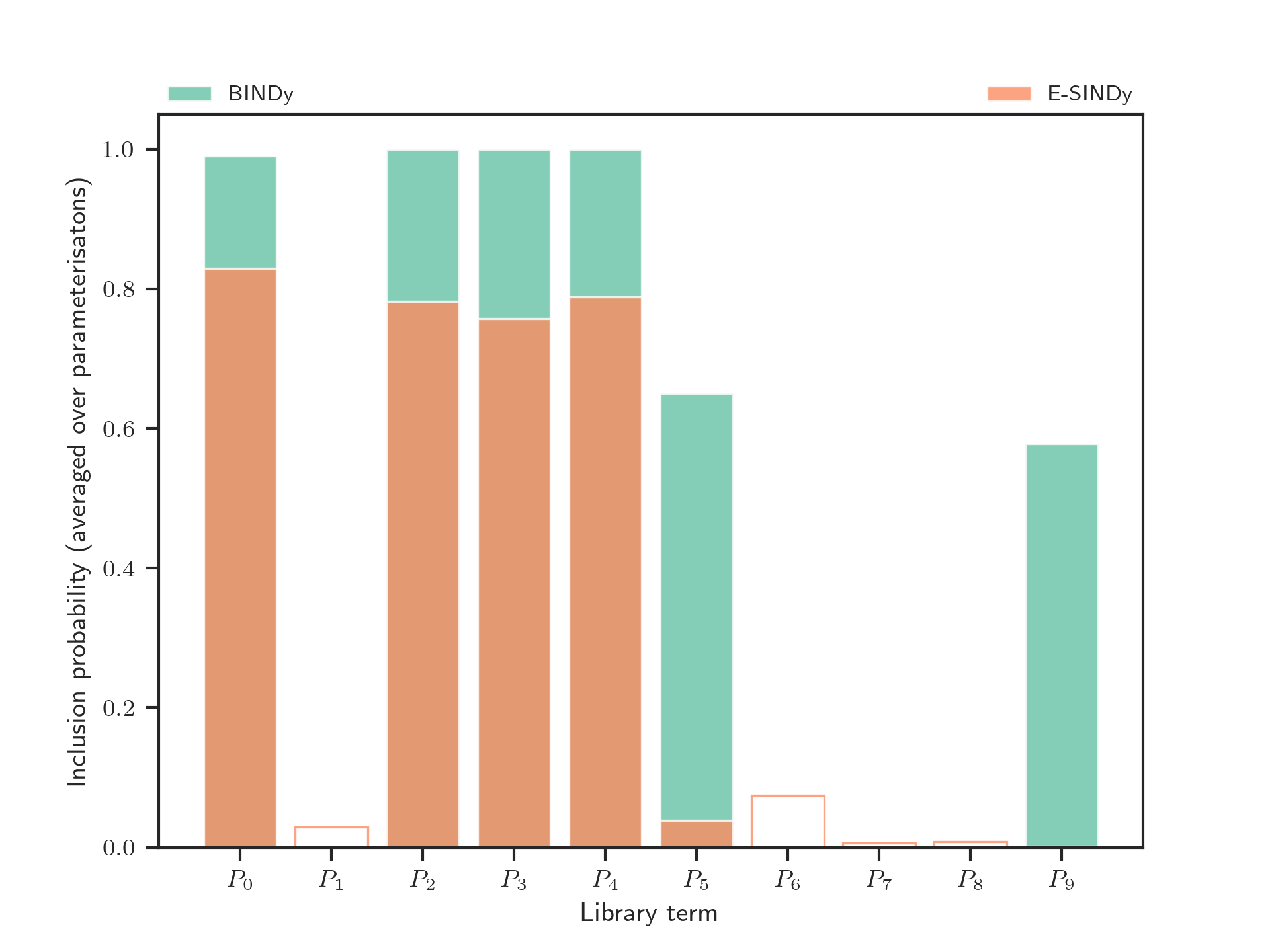}
	\caption{\rev{Posterior probabilities of BINDy and E-SINDy of selecting polynomial terms, averaged over 100 parametrisations of the Legendre polynomial in the first case study. Solid blocks correspond to terms that are truly in the data-generating model, hollow bars correspond to spurious terms.}}
	\label{fig:poly_param}
\end{figure}

\rev{The results of this investigation are plotted in Figure \ref{fig:poly_param}. As can be seen in the figure, BINDy assigns higher probability to the correct model terms in every case. It is interesting to note that the small-valued parameters ($P_5$ and $P_9$) are only identified by BINDy at around 60\% of the time on average, less confidence than was seen in the specific parameterisation of the results in figure \ref{fig:poly_post}. Some of this drop might be explained by the fact that sampling parameters in [0,1], shrinking them by a factor of 100 and then rounding to three decimal places will almost certainly produce some parameters that have zero magnitude or a magnitude so small as to be indistinguishable from noise. However, the authors would note that even this slightly reduced performance represents a significant improvement over the E-SINDy approach that fails to identify the small-valued parameters in almost every case.}

\section{\rev{Convergence of BINDy}}

\begin{figure}
	\captionsetup{labelfont={color=\revcolorflag}}
	\centering
	\includegraphics[width=\fw]{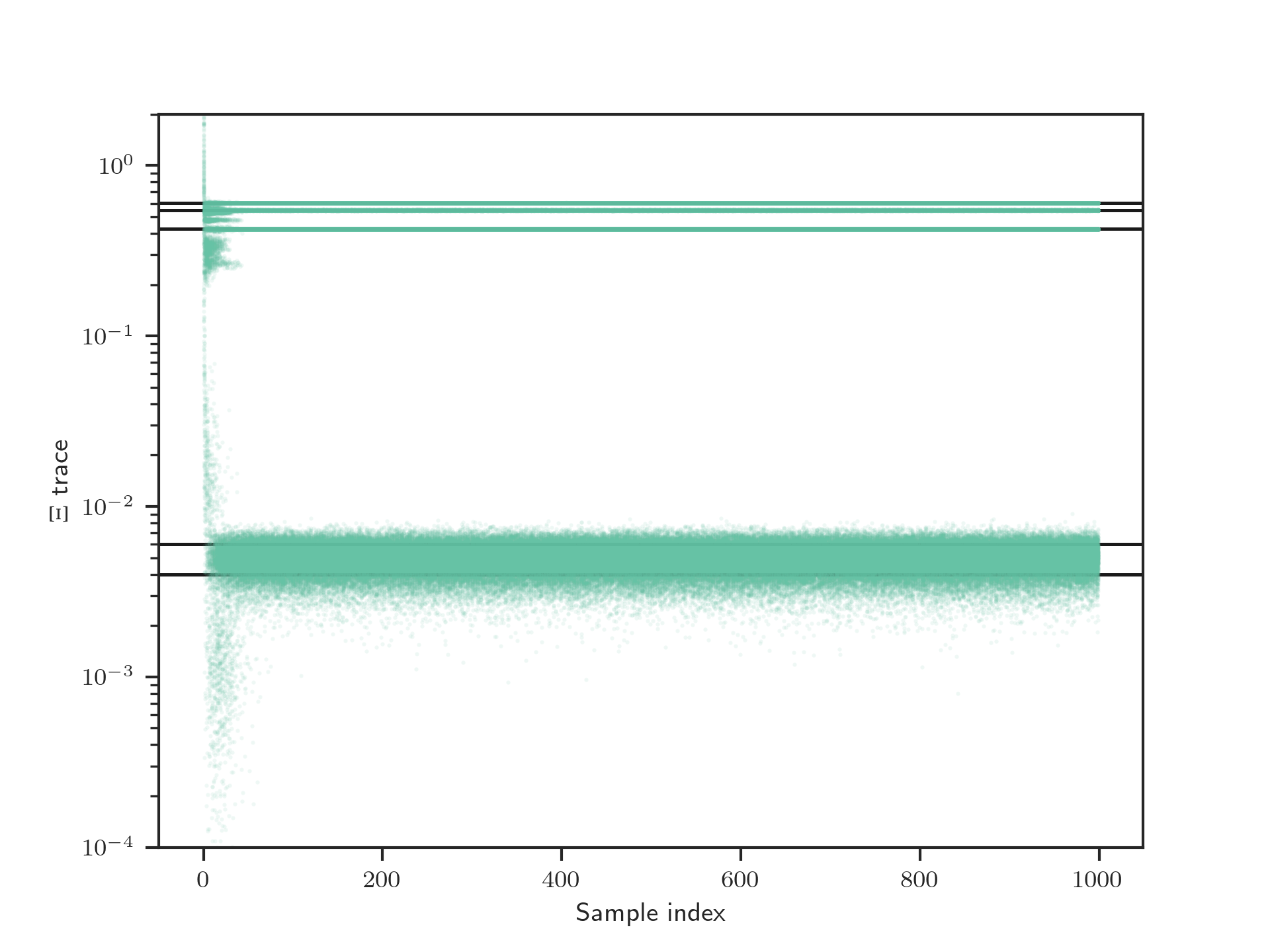}
	\caption{\rev{MCMC Trace of the elements of $\Xi$ (fitted to the polynomial case-study). Superimposed trace of 100 chains with initial conditions drawn from the prior.}}
	\label{fig:trace}
\end{figure}

\begin{figure}
	\captionsetup{labelfont={color=\revcolorflag}}
	\centering
	\includegraphics[width=\fw]{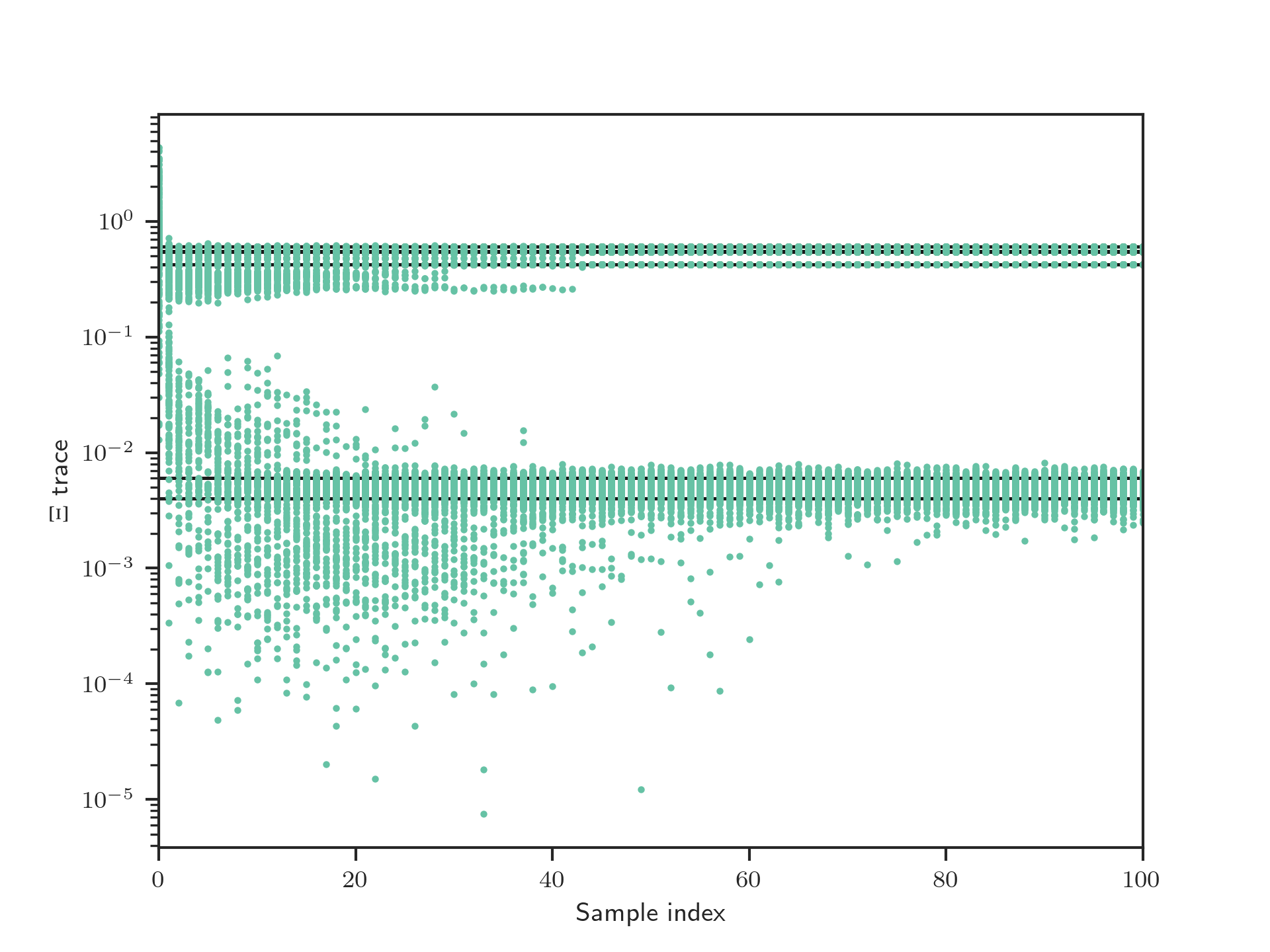}
	\caption{\rev{Zoomed view of traces of the elements of $\Xi$ for the first 100 iterations of the sampler.}}
	\label{fig:trace_zoom}
\end{figure}

\rev{The convergence of RJMCMC methods depends strongly on the choices of the prior distribution and jump kernel as well as the geometry of the posterior. Here we consider the convergence of the proposed scheme to the polynomial case study example in the main body of the paper. In total 100 samples are drawn from the prior (as defined above) as initial conditions for the Gibbs sampler. The sampler is then run for 1000 iterations, with no samples discarded so that the convergence during the burn in period is visible. The trace of the parameter values in $\Xi$ is plotted in Figures \ref{fig:trace} and \ref{fig:trace_zoom}.}

\rev{As can be seen in Figure \ref{fig:trace}, the values of the parameters (green dots) converge quickly to the true values (horizontal black lines) and remain there. Although there is increased variance for the small valued parameters (corresponding to $P_5$ and $P_9$ in the polynomial library), it appears visually to be stationary indicating that the chain has converged. Figure \ref{fig:trace_zoom} depicts a zoom on the first 100 iterations of the sampler. Here it can be seen that despite high initial variance, the chain quickly converges and appears well-converged after only 100 iterations.}

\rev{Although the simplistic 'bit-flipping' jump kernel used in this study has a number of theoretical limitations, it appears that it is sufficient to produce a well converged chain in this example.}

\section{\rev{Robustness of BINDy}}

\rev{An important concept in SINDy-type modelling is the robustness in the high-noise, low-data regime. Indeed, this is a practical consideration for nonlinear system identification approaches more generally. Most SINDy-type approaches return point estimates of the model and its parameterisation, given the cognitive bias of sparsity injected into the chosen sparse regression scheme. In this context, one can interpret robustness as choosing the `correct' data-generating model.}

\rev{The proposed approach targets the posterior probability distribution $p(\Xi, m | \dot{x}, \Theta(x))$. This distribution explicitly incorporates sources of uncertainty such as low-data and high-noise by way of the data likelihood model in \eqref{eqn:sindy_bayes}. Thus, the probability of selecting the correct model is very likely to diminish in the presence of high noise or low data (as the data likelihood would become more diffuse). For even very large Gaussian noise, with perfect state observation the conjugate update of the parameters is exact (for a given model order) and so the posterior would exactly reflect the uncertainty due to noise. However, in practice, large noise levels are very likely to introduce non-Gaussian disturbances to the state derivative computation (as is the case in SINDy methods generally) and so this would likely be the limiting factor in practice.}

\begin{figure}
	\captionsetup{labelfont={color=\revcolorflag}}
	\centering
	\includegraphics[width=\fw]{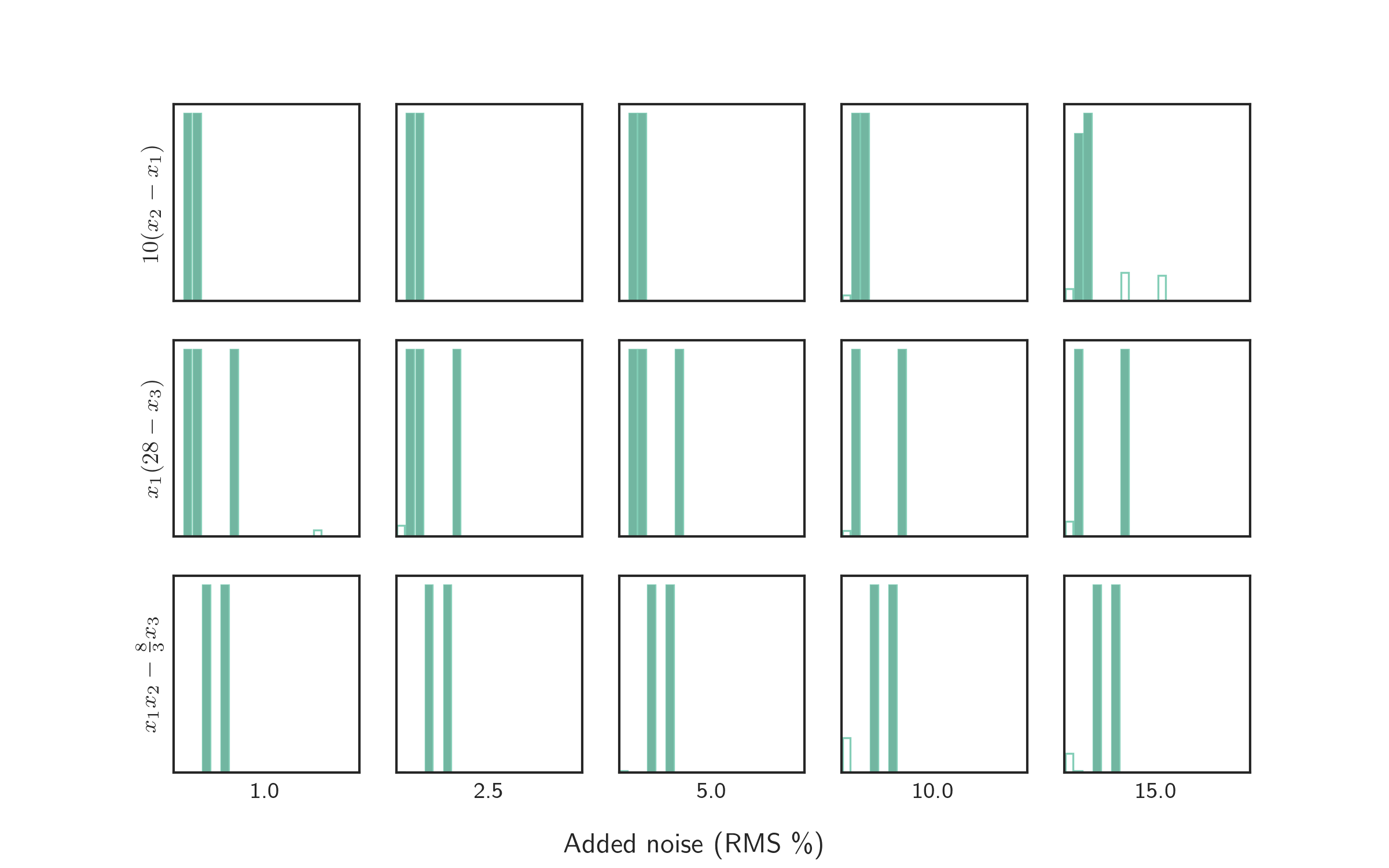}
	\caption{\rev{Posterior term inclusion probabilities at a fixed data length for various levels of added Gaussian noise. Estimated by BINDy with a flat model prior. Subplot axes are identical to those of Figure \ref{fig:lorz_post}. Solid bars indicate terms truly in the data generating model, while hollow bars indicate spurious terms.}}
	\label{fig:const_len}
\end{figure}

\begin{figure}
	\captionsetup{labelfont={color=\revcolorflag}}
	\centering
	\includegraphics[width=\fw]{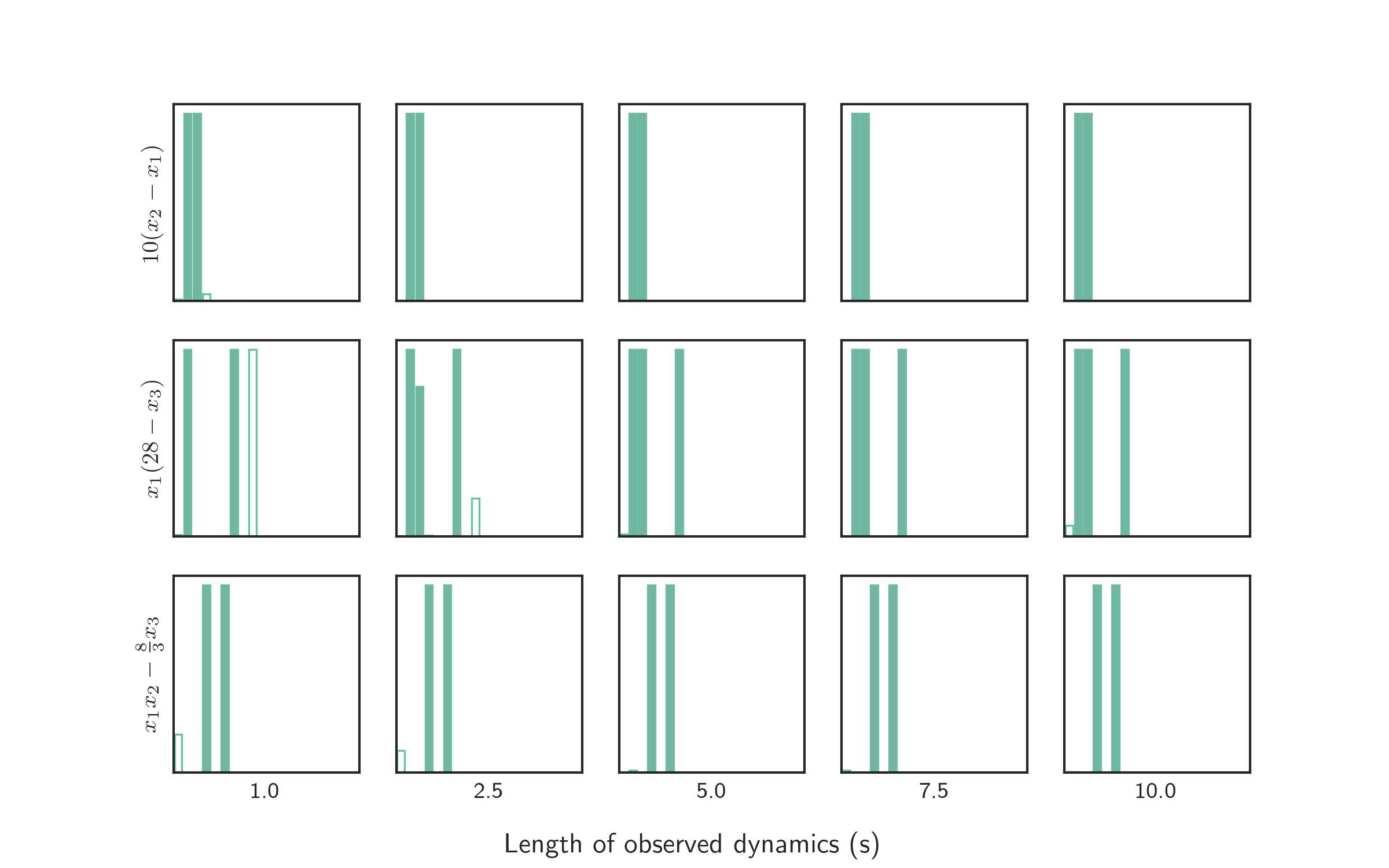}
	\caption{\rev{Posterior term inclusion probabilities at a fixed noise level for various amounts of observed dynamics. Estimated by BINDy with a flat model prior. Subplot axes are identical to those of Figure \ref{fig:lorz_post}. Solid bars indicate terms truly in the data generating model, while hollow bars indicate spurious terms.}}
	\label{fig:const_noise}
\end{figure}

Nevertheless, it is instructive to consider the effect of low-data and high-noise regimes. In Figures \ref{fig:const_len} and \ref{fig:const_noise}, the posterior term inclusion probabilities are plotted. BINDy is applied to the Lorenz case-study example with a flat model prior. In Figure \ref{fig:const_len} the length of the observed data is held constant at 10s (or 1000 observations) and the variance of the added noise is varied from 1\% to 15\%. As can be seen in the Figure, BINDy is very robust to noise and assigns high posterior probability to correct terms in the data-generating model, with some spurious and missed terms assigned low probability at the highest noise levels.

In Figure \ref{fig:const_noise}, the effect of decreasing amounts of available data is investigated. The level of added noise is held constant at 2.5\% RMS and the length of the observed dynamics available for inference is varied between 1.0 seconds (corresponding to only 100 datapoints) up to 10 seconds (corresponding to 1000 datapoints). The proposed BINDy approach with a flat model prior is applied to the data. Once again BINDy performs excellently, correctly assigning high probability to terms in the Lorenz equations with some spurious and missed terms in the shortest data lengths. 

\begin{figure}
	\captionsetup{labelfont={color=\revcolorflag}}
	\centering
	\includegraphics[width=0.8\linewidth]{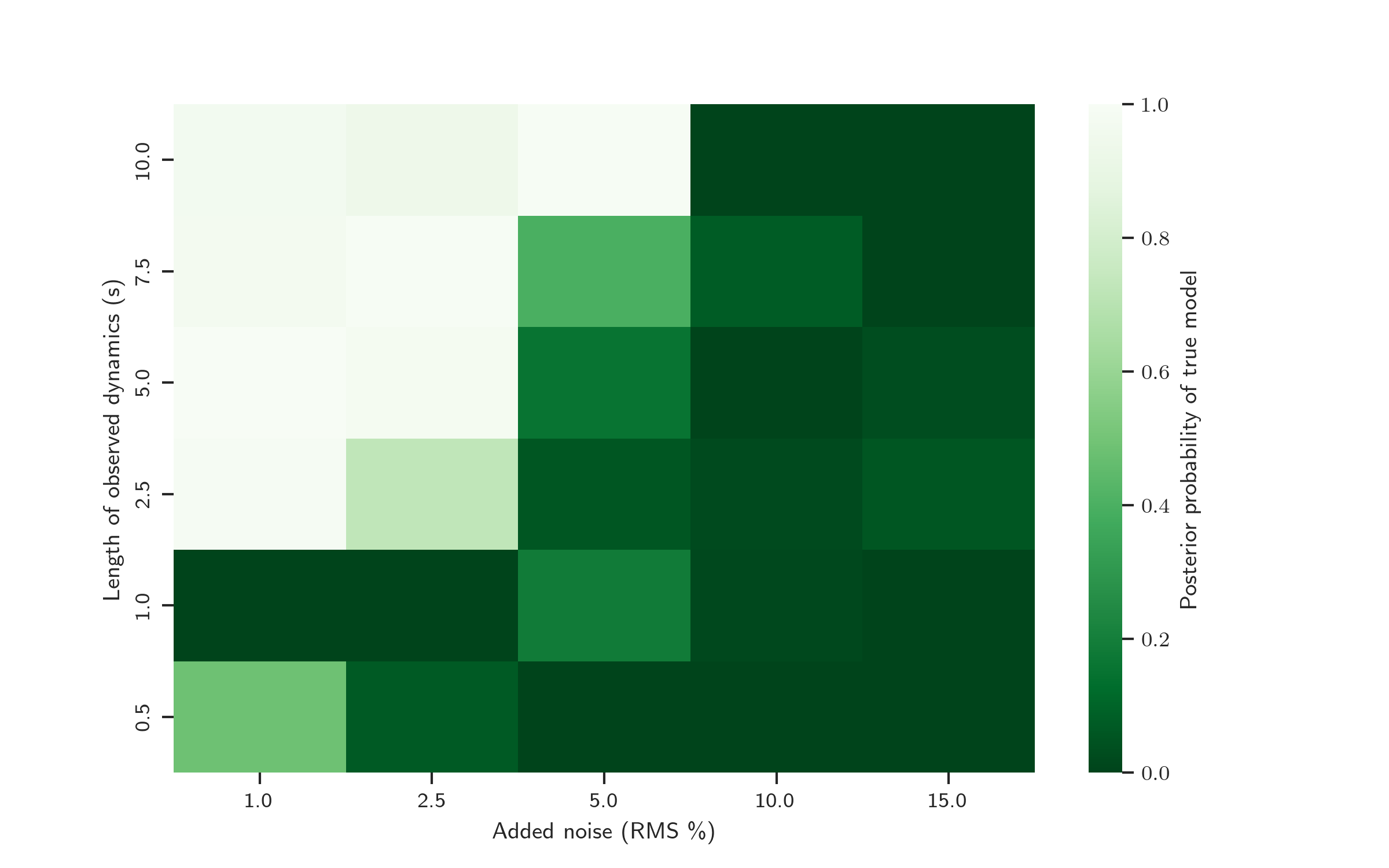}
	\caption{\rev{Heatmap depicting the posterior probability of the true model for different noise levels and data lengths.}}
	\label{fig:heatmap}
\end{figure}

\rev{Figure \ref{fig:heatmap} depicts the robustness of BINDy to the effect of data length and noise in a heatmap format. Each element of the heatmap considers the probability of the true data-generating model under the posterior estimated by BINDy. As can be seen in the figure, the probability approaches unity for low noise and long sequences.}

\end{document}